%% file: main.tex
\documentclass[10pt,twocolumn,letterpaper]{article}
\usepackage[pagenumbers]{cvpr}

\input{preamble}
\usepackage{booktabs}
\usepackage{graphicx}
\usepackage{colortbl}

\definecolor{ihmrblue}{HTML}{0A66D8}
\definecolor{ihmrzone}{HTML}{E3EEFB}
\definecolor{ihmrgrey}{HTML}{7C7B77}
\usepackage{multirow}
\usepackage{xspace}
\newcommand{\method}{InstantHMR\xspace}

\definecolor{linkblue}{rgb}{0.21,0.49,0.74}
\usepackage[pagebackref,breaklinks,colorlinks,allcolors=linkblue]{hyperref}

\input{sec/supp_sections}

\title{\input{sec/title}}

\author{
Mohamed Adjel\textsuperscript{1} \qquad Kahina Chalabi\textsuperscript{2} \qquad Vincent Bonnet\textsuperscript{2}\\
\textsuperscript{1}NaturalPad, Montpellier, France \qquad \textsuperscript{2}LAAS-CNRS, Toulouse, France\\
{\tt\small mohamed.adjel@naturalpad.fr, \{kchalabi, vbonnet\}@laas.fr}
}

\begin{document}
\maketitle
\input{sec/0_abstract}
\input{sec/1_intro}
\input{sec/2_related}
\input{sec/3_method}
\input{sec/4_protocol}
\input{sec/5_experiments}
\input{sec/6_conclusion}
\section*{Acknowledgements}
This project was provided with computer and storage resources by GENCI at IDRIS thanks to the grant 2025-AD011017116 on the V100 partition of the supercomputer Jean Zay.
We thank Meta for releasing the annotations of the SAM~3D~Body dataset and the MHR body model.

\input{sec/llm_statement}
{
    \small
    \bibliographystyle{ieeenat_fullname}
    \bibliography{main}
}

\clearpage
\onecolumn
\appendix
\begin{center}
  {\Large\bfseries \input{sec/title}\par}
  \vspace{0.5em}
  {\large Supplementary Material\par}
\end{center}
\vspace{1em}

\input{sec/supp_body}
\end{document}

%% file: sec/supp_sections.tex
\newcommand{\suppNetwork}{\ref{sec:supp-network}}
\newcommand{\suppLosses}{\ref{sec:supp-losses}}
\newcommand{\suppReproduction}{\ref{sec:supp-as-published}}

\newcommand{\suppHistory}{\ref{sec:supp-history}}
\newcommand{\suppFit}{\ref{sec:supp-fit}}
\newcommand{\suppTemporal}{\ref{sec:supp-temporal}}
\newcommand{\suppPareto}{\ref{sec:supp-pareto}}
\newcommand{\suppPartial}{\ref{sec:supp-partial}}
\newcommand{\suppDevice}{\ref{sec:supp-device}}

%% file: sec/title.tex
InstantHMR: Real-Time Human Mesh Recovery on Your Phone

%% file: sec/0_abstract.tex
\begin{abstract}
Real-time 3D human mesh recovery on phones serves robotics, biomechanics, AR/VR and exergames, but current models are either too slow for mobile devices or sacrifice accuracy for speed.
We present \method, a family of open-source single-image regressors built for phone NPUs. \method predicts the Momentum Human Rig (MHR) with a convolutional backbone and a small transformer decoder, and is trained only on the publicly released annotations of the SAM~3D~Body dataset.
On a Samsung Galaxy S23 NPU, \method-M recovers a 6,890-vertex mesh in 5.5\,ms per person (0.98\,ms on an RTX~4070 GPU), faster than every method we evaluate, at 48.8\,mm PA-MPJPE on EMDB, 40.0\,mm on 3DPW and 41.7\,mm on RICH.
An optional kinematic fit, a single optimisation step inside the exported graph that aligns the body with the network's own 2D keypoints, matches NLF-S on EMDB and 3DPW in 6.4\,ms on the phone, 10.6$\times$ faster than NLF-S.
For fair comparisons, we re-evaluate 11 methods under one protocol matched to the closest MHR-based competitors, with Meta's official MHR-to-SMPL conversion, and measure latency on eight targets, from desktop TensorRT to two phone NPUs.
Code and models are available at \url{https://github.com/Naturalpad-studio/InstantHMR}.
\end{abstract}

%% file: sec/1_intro.tex
\section{Introduction}
\label{sec:intro}

\begin{figure}[t]
  \centering
  \includegraphics[width=\columnwidth]{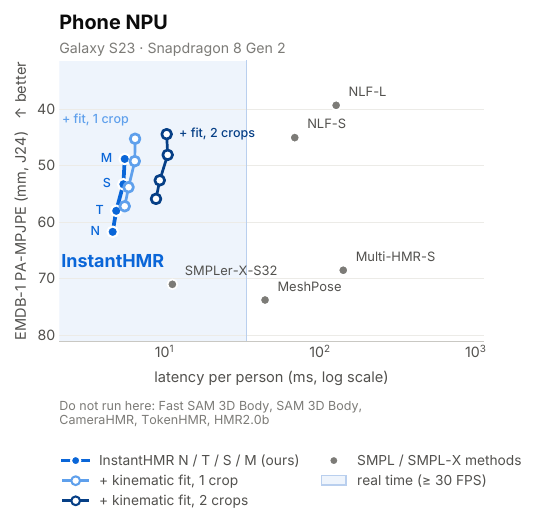}
  \caption{\textbf{Accuracy against latency on a phone.} EMDB-1 PA-MPJPE (24 joints, lower is better) against latency per person on the Galaxy S23 NPU (log scale), for every method under the protocol of \cref{sec:protocol}. Blue: \method's four sizes, as the network alone (filled circles) and with the kinematic fit of \cref{sec:method-fit} on one crop (light) or two crops (dark). \colorbox{ihmrzone}{Shaded}: real time ($\geq$\,30\,FPS). HybrIK-R34 is not shown: its inverse kinematics does not export, so it cannot be timed whole on the phone (supplementary material, Sec.~\suppPartial). Desktop GPU and 3DPW: supplementary material, Sec.~\suppPareto.}
  \label{fig:teaser}
\end{figure}

Monocular real-time human mesh recovery (HMR) is a long-standing goal in computer vision, with applications in robotics~\cite{he2024h2o}, tele-rehabilitation~\cite{uhlrich2023opencap}, workplace ergonomics~\cite{li2020rula,jiao2024reba}, AR/VR~\cite{mehta2017vnect,jiang2022avatarposer}, exergames~\cite{zaoui2026rhythmic}, and 3D animation~\cite{shimada2020physcap}.
Since 2023, most smartphones include a neural processing unit (NPU), which is a formidable opportunity for local real-time inference, fostering privacy, low energy consumption and ease of use.

Recent models recover a body mesh from a single image with remarkable robustness~\cite{sarandi2024nlf,patel2025camerahmr,yang2026sam3dbody}, but they are built for GPUs: SAM~3D~Body~\cite{yang2026sam3dbody} takes 416\,ms per person on a desktop GPU in its own release and does not run on a phone's NPU at all. 
NLF ~\cite{sarandi2024nlf} is the fastest state-of-the-art model on a desktop GPU (1.52\,ms per person for the smallest variant), but is not tailored for real-time mobile inference (67.97\,ms per person on a Samsung Galaxy S23 NPU). 
On the other hand, models that run in real time on a phone (under 33\,ms per frame, \ie 30\,FPS), such as SMPLer-X-S32~\cite{cai2023smplerx} on the S23 NPU or MeshPose on an iPhone~12, as reported by its authors~\cite{le2024meshpose}, give up much of the accuracy (cf.\ \cref{tab:accuracy} and \cref{fig:teaser}).

We present \method, a family of single-image HMR models designed for phones.
\method regresses the parameters of the Momentum Human Rig (MHR)~\cite{ferguson2025mhr} with an efficient convolutional backbone \cite{zhao2024rtdetr} and a small transformer decoder. It is trained on the publicly released annotations of SAM~3D~Body's training set.
On a Galaxy S23 NPU, \method-M recovers a 6,890-vertex mesh in 5.5\,ms per person, 12$\times$ and 23$\times$ faster than NLF-S and NLF-L, the only phone-compatible methods that are more accurate (\cref{fig:teaser,tab:speed}).
An optional kinematic fit, exported in the same graph, makes the 3D body agree with the network's own 2D keypoints, which are more accurate than the projection of its 3D output; with it, \method-M matches NLF-S on EMDB-1 and 3DPW at 10.6$\times$ its speed on the phone.
Published numbers on EMDB~\cite{kaufmann2023emdb} and 3DPW~\cite{vonmarcard2018pw3d} are produced under different person boxes, focal lengths, joint sets and conversions between body models, and these choices move a single model by several millimetres, as much as the gaps between methods.
We therefore re-evaluate eleven methods under one protocol. Our contributions are:
\begin{itemize}
  \item \textbf{\method}, a family of four MHR regressors (nano to medium) trained only on released annotations; the medium size is the most accurate method that runs in real time on a phone NPU.
  \item \textbf{A kinematic fit inside the deployed graph}: one Levenberg-Marquardt step that moves the body pose until its projection matches the network's own 2D keypoints, with no learned parameters, running in fp16 entirely on the phone NPU.
  \item \textbf{A matched evaluation protocol} with Meta's official MHR-to-SMPL conversion, under which eleven competitors are re-scored and, when their paper reports the released variant, checked against their published numbers.
  \item \textbf{Latency on eight targets}, from desktop TensorRT to two phone NPUs and two phone CPUs, measured the same way for every method, and end-to-end timings in an Android app.
\end{itemize}
The code (Apache-2.0) and the models (for non-commercial use) are available at \url{https://github.com/Naturalpad-studio/InstantHMR}, with a variant trained only on commercially usable images (\method-M-C, \cref{tab:family}).

%% file: sec/2_related.tex
\section{Related work}
\label{sec:related}

\begin{figure*}[t]
  \centering
  \includegraphics[width=\textwidth]{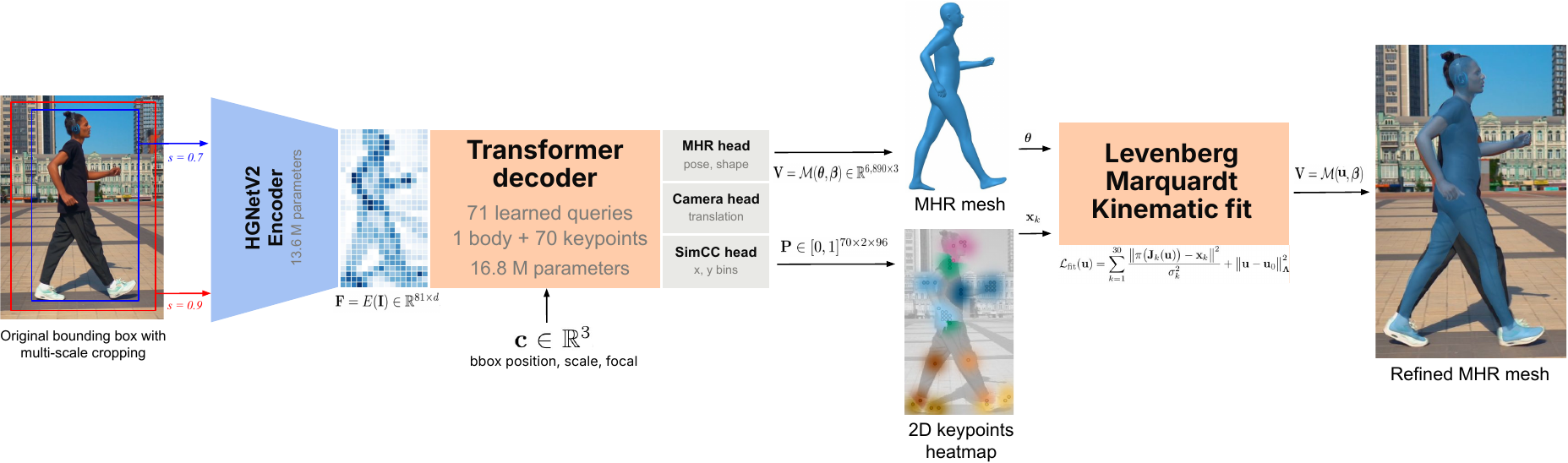}
  \vspace{-1.2em}
  \caption{\textbf{\method.} A 288\,px crop of the person box goes through a convolutional HGNetV2 encoder down to $9\times9$ image tokens. A 4-layer transformer decoder refines 71 learned queries (one for the body, one per keypoint), to which the embedded position of the box in the full frame, $\mathbf{c}$, is added. The body query feeds the MHR and camera heads, and the rig gives the posed mesh; each keypoint query feeds a SimCC head that classifies its $x$ and $y$ over 96 bins. The optional kinematic fit (\cref{sec:method-fit}) takes one Levenberg-Marquardt step that moves the body until its projection matches these 2D keypoints. For the fit, the box is shrunk by $s=0.9$, and by $s=0.7$ for an optional second crop. Every visual is a real output of \method-M on a 3DPW test frame~\cite{vonmarcard2018pw3d}.}
  \label{fig:overview}
\end{figure*}

\paragraph{Parametric body models.}
Most HMR methods predict the parameters of a body model rather than a mesh.
SMPL~\cite{loper2015smpl}, with 6,890 vertices and 24 joints, is the most used; SMPL-X~\cite{pavlakos2019smplx} adds hands and face, MHR~\cite{ferguson2025mhr} decouples the skeleton from the body shape, Anny~\cite{bregier2025anny} replaces the scan-based shape space with interpretable parameters, and SOMA~\cite{saito2026soma} maps these models to one canonical mesh.
They differ in topology, skeleton and units, so comparing methods that predict different models requires converting between them (\cref{sec:protocol}); \method predicts MHR.

\paragraph{Human mesh recovery from a single image.}
HMR~\cite{kanazawa2018hmr} regresses SMPL parameters from a single image.
HMR2.0~\cite{goel2023hmr2} scales this formulation with a ViT backbone and large pseudo-labelled data, TokenHMR~\cite{dwivedi2024tokenhmr} predicts the pose as discrete tokens, CLIFF~\cite{li2022cliff} conditions the regressor on the location of the person in the full frame, and CameraHMR~\cite{patel2025camerahmr} adds a camera-aware head trained with estimated focal lengths.
HybrIK~\cite{li2021hybrik} recovers the rotations from 3D keypoints with analytical inverse kinematics, and NLF~\cite{sarandi2024nlf} predicts any body point as a continuous field, which allows training one model on annotations from different skeletons and meshes.
SMPLer-X~\cite{cai2023smplerx} and Multi-HMR~\cite{baradel2024multihmr} regress SMPL-X, for one person or for many at once.
SAM~3D~Body~\cite{yang2026sam3dbody} moves to MHR with a promptable DINOv3-based architecture trained on a large human-annotated corpus, Fast SAM~3D~Body~\cite{fastsam3dbody2026} accelerates it by up to 10.9$\times$ without retraining, and GEM-X~\cite{nvidia2026gemx} also builds on DINOv3 and predicts SOMA.
These models are remarkably robust, but they are built for GPUs.

\paragraph{Fitting to 2D keypoints.}
Optimisation-based methods such as SMPLify~\cite{bogo2016smplify} fit a body model to detected 2D keypoints, by minimising their reprojection error under pose and shape priors over many iterations.
SPIN~\cite{kolotouros2019spin} runs such a fit inside the training loop, EFT~\cite{joo2021eft} fine-tunes the regressor on each test image, and later work learns the update steps~\cite{song2020lgd} or guides a diffusion model with the 2D evidence~\cite{stathopoulos2024scorehmr}.
PyMAF~\cite{zhang2021pymaf} and ReFit~\cite{wang2023refit} instead feed the alignment between the mesh and the image back into the network, over several learned refinement steps.
Our fit is simpler: a single step with no learned parameters, which uses the network's own 2D keypoints rather than an external detector and runs inside the exported graph on the phone (\cref{sec:method-fit}).

\paragraph{Video and on-device recovery.}
Video methods use the temporal context to recover smoother motion, often in world coordinates: WHAM~\cite{shin2024wham} runs online from past frames, TRAM~\cite{wang2024tram}, GVHMR~\cite{shen2024gvhmr} and GEM~\cite{li2025gem} process whole sequences, and the video model of GEM-X~\cite{nvidia2026gemx} follows the motion from the features of its single-image model.
Many applications, such as AR/VR, tele-rehabilitation or robot teleoperation, need the pose of a frame while that frame is captured, and a method that waits for future frames adds latency: FastHMR~\cite{mehraban2026fasthmr} reaches 103--150\,FPS on an RTX~3090, but waits for up to 242 later frames, about 8\,s at 30\,FPS.
Online methods such as WHAM avoid this wait, but still need a per-frame image model: its reported 200\,FPS exclude its keypoint detector and image encoder~\cite{shin2024wham}.
We therefore focus on the single-image model, the per-frame building block of online video methods.
Efficient single-image HMR has mostly targeted GPUs: Seeing Machines' HMR model runs in 5.5--15.8\,ms on an NVIDIA Jetson Thor robotics module~\cite{seeingmachines2026hmr}.
On phones, BlazePose GHUM Holistic~\cite{grishchenko2022blazeposeghum} lifts real-time 3D landmarks to a body model, but is not evaluated on a standard mesh benchmark.
The closest work to ours, MeshPose~\cite{le2024meshpose}, localises a low-poly SMPL mesh with DensePose-style supervision at 97--153\,FPS on an iPhone~12; we evaluate its released model in \cref{sec:experiments}.

\paragraph{Evaluation protocols.}
3DPW~\cite{vonmarcard2018pw3d}, EMDB~\cite{kaufmann2023emdb} and RICH~\cite{huang2022rich} are the standard in-the-wild benchmarks, but papers score them with different person boxes, focal lengths and conversions to SMPL: SAM~3D~Body uses Meta's iterative MHR-to-SMPL fitter, Fast SAM~3D~Body a learned network, and GEM-X its own dataset-specific projectors.
We fix one protocol and re-run every method under it (\cref{sec:protocol}).

%% file: sec/3_method.tex
\section{\method}
\label{sec:method}

\method maps a square crop of the person, $\mathbf{I}\in\mathbb{R}^{288\times288\times3}$, and the position of that crop in the full image, $\mathbf{c}\in\mathbb{R}^{3}$, to three outputs (\cref{fig:overview}): the MHR rig parameters $\boldsymbol{\theta}\in\mathbb{R}^{204}$ (root translation and rotation, 130 joint parameters and 68 bone scales) and identity coefficients $\boldsymbol{\beta}\in\mathbb{R}^{45}$; 70 2D keypoints $\mathbf{x}\in\mathbb{R}^{70\times2}$ in crop coordinates; and the camera translation $\mathbf{t}\in\mathbb{R}^{3}$.
The MHR rig $\mathcal{M}$ turns $(\boldsymbol{\theta},\boldsymbol{\beta})$ into the posed mesh $\mathbf{V}\in\mathbb{R}^{6{,}890\times3}$, a farthest-point subset of MHR's 18,439 vertices with as many vertices as SMPL, and its 70 3D keypoints $\mathbf{J}$.
Every design choice serves one constraint: the whole network must compile to a phone NPU, in fp16, with no operator falling back to the CPU. The equations are in Secs.~\suppNetwork{} and \suppLosses{} of the supplementary material.

\subsection{Network}
\paragraph{HGNetV2 Backbone.}
A convolutional HGNetV2~\cite{zhao2024rtdetr} backbone $E$, pretrained on ImageNet, turns the $288\times288$ crop into a $9\times9$ grid of features.
A linear layer projects them to $d=512$ channels, and a fixed 2D sine-cosine positional embedding is added, which gives the 81 image tokens $\mathbf{F}=E(\mathbf{I})\in\mathbb{R}^{81\times d}$.
Convolutions, unlike the attention of a ViT backbone, map directly onto the operators phone NPUs accelerate.
The four sizes of \method differ only in the backbone (B0, B1, B2, B4 for nano, tiny, small, medium), and have 20.0, 20.3, 24.4 and 32.2\,M parameters.
The decoder and the heads are the same in every size, and the decoder alone holds 16.8\,M of these parameters.

\paragraph{CLIFF conditioning.}
\label{sec:method-position}
As in CLIFF~\cite{li2022cliff}, the network receives the position of the crop in the full image, since the same crop could come from the centre or the edge of a wide-angle image.
We express it through the focal length $f$, so that it keeps one meaning across cameras: $\mathbf{c}$ holds the two viewing angles of the box centre and the box size over $f$, and a two-layer MLP $\phi$ embeds it into $\phi(\mathbf{c})\in\mathbb{R}^{d}$.
Without calibration, $f$ defaults to $1.05\times$ the image diagonal.

\paragraph{Transformer Decoder.}
71 learned queries $\mathbf{Q}\in\mathbb{R}^{71\times d}$, one for the body and one per 2D keypoint, receive the position embedding $\phi(\mathbf{c})$, added to every query.
A 4-layer pre-norm transformer decoder~\cite{carion2020detr} $D$ (8 heads) then lets them attend to each other and to the image tokens $\mathbf{F}$, and outputs one feature $\mathbf{z}_i\in\mathbb{R}^{d}$ per query.
The feature $\mathbf{z}_0$ comes from the body query, and $\mathbf{z}_1,\dots,\mathbf{z}_{70}$ from the keypoint queries.

\paragraph{Decoder Heads.}
Three heads read the decoder output.
The SimCC head $h_{\mathrm{2D}}$, a two-layer MLP shared by all keypoints, turns each keypoint query into its two distributions, $\mathbf{P}_k = h_{\mathrm{2D}}(\mathbf{z}_k)$, whose expectations over the bin centres give the 2D keypoint $\mathbf{x}_k$.
Classifying the coordinates rather than regressing them keeps the image evidence of every joint in the decoder.
The camera head $h_{\mathrm{cam}}$, a linear layer, predicts the camera translation from the body query, $\mathbf{t} = h_{\mathrm{cam}}(\mathbf{z}_0)$.
The MHR head $h_{\mathrm{MHR}}$ predicts the body from the same query, $(\boldsymbol{\theta},\boldsymbol{\beta}) = h_{\mathrm{MHR}}(\mathbf{z}_0)$.
It uses SAM~3D~Body's continuous parameterisation (a 6D root rotation~\cite{zhou2019continuity}, the continuous body pose and bone-scale principal components), with the bone scales bounded to the range seen in training, which keeps the training stable (supplementary material, Sec.~\suppNetwork).
The 3D keypoints are not regressed: they are read from the posed rig through its own forward kinematics, so 3D keypoints and mesh can never disagree.

\begin{table*}[t]
  \centering
  \small
  \setlength{\tabcolsep}{4.5pt}
  \resizebox{\textwidth}{!}{\input{tables/tab_speed}}
  \caption{\textbf{Latency on eight targets.} Latency per person in ms (batch 1, from the crop to a 6,890-vertex mesh), with EMDB-1 PA-MPJPE in mm for context. Desktop: RTX~4070 (TensorRT fp16, ONNX Runtime fp32, each method's PyTorch release) and Intel i5-14400F CPU (ONNX Runtime). Phones: Galaxy S23 and S24 NPUs (fp16), S23 and A14 CPUs (fp32), through Qualcomm AI Hub. \textcolor{ihmrblue}{Blue}: ours. \textbf{Bold}: best; \underline{underlined}: best on a phone NPU. \colorbox{ihmrzone}{Shaded}: real time on the phone ($\leq$\,33.3\,ms). Grey factors: times slower than \method-M. $\times$: does not run on the device. --: cannot be timed as a whole method (partial timings, as lower bounds, in the supplementary material, Sec.~\suppPartial). \textbf{+ fit}: \cref{sec:method-fit}.}
  \label{tab:speed}
\end{table*}

\begin{table*}[t]
  \centering
  \small
  \resizebox{\textwidth}{!}{\input{tables/tab_accuracy}}
  \caption{\textbf{Accuracy under one protocol} (\cref{sec:protocol}). Errors in mm on EMDB-1 and 3DPW (every 10th frame) and on the 21,287 RICH test crops; latency per person in ms. \textbf{+ fit}: the kinematic fit of \cref{sec:method-fit}, on one or two crops. \textcolor{ihmrblue}{Blue}: ours. \textbf{Bold}: best; \underline{underlined}: best on a phone NPU. \colorbox{ihmrzone}{Shaded}: real time on the phone. Grey factors: times slower than \method-M. MeshPose has no metric scale, so only PA-MPJPE is defined, and it is not scored on RICH. $^\dagger$, $^\ddagger$: trained on 3DPW-train, on RICH. $^p$: PyTorch release; $^t$: TensorRT backbone + PyTorch.}
  \label{tab:accuracy}
\end{table*}

\subsection{Training}
\paragraph{Dataset.}
\method is trained on the human-annotated MHR fits that Meta released with SAM~3D~Body~\cite{yang2026sam3dbody}, over the images of five of its sources: SA-1B~\cite{kirillov2023sam}, AI Challenger~\cite{wu2017aichallenger}, Harmony4D~\cite{khirodkar2024harmony4d}, COCO~\cite{lin2014coco} and MPII~\cite{andriluka2014mpii}, 4.2\,M person crops in total, sampled at 43\,/\,26\,/\,14\,/\,13\,/\,4\,\%.
No 3DPW or EMDB images are used.
The labels are the dataset's annotations, not the predictions of a teacher model, so the student is not capped at a teacher's accuracy.

\paragraph{Losses.}
The training loss is a weighted sum of eleven terms (supplementary material, Sec.~\suppLosses).
Five compare the predicted MHR parameters and camera translation with the annotated ones.
Three compare the geometry that the rig produces from them: the 70 3D keypoints, 595 vertices spread over the mesh, and the 127 skeleton joints.
Three supervise the image side: a SimCC cross-entropy against a Gaussian label on each keypoint, a distance on the 2D keypoints, and the reprojection of the 3D keypoints with the calibrated camera.
As in NLF~\cite{sarandi2024nlf}, 3D errors are unsquared Euclidean distances, so that a few bad samples do not dominate the gradient.
The vertex term matters for shape: $\boldsymbol{\beta}$ moves the mesh but not the skeleton, so without it, it receives no geometric gradient.

\paragraph{Training Recipe and Deployment.}
Each model is trained for 600 epochs of 400\,k samples, batch 256 on four GPUs, with AdamW and a one-cycle learning-rate schedule peaking at $6\times10^{-4}$.
Each training crop is either the standard crop (probability 0.75) or one of three other framings cut from the full image (zoomed out, zoomed in, shifted), with synthetic occlusion. This mix is the largest single gain we measured: against training on the standard crop alone, it improves EMDB-1 PA-MPJPE by 10.0\,mm (58.8 to 48.8) and 3DPW by 2.7\,mm, more than an even or uniform mix of the framings. The 288\,px input is the second: at every backbone size, it improves 3DPW validation PA-MPJPE by 1.8--4.0\,mm over 224\,px at the same number of epochs (supplementary material, Sec.~\suppHistory).

The network exports to a single ONNX graph, which runs in fp16 on the S23 NPU with all of its 1,607 layers on the NPU.
For the latency we report, the graph also includes the rig's forward kinematics and linear blend skinning up to the posed mesh $\mathbf{V}$.

\subsection{Kinematic fit to the 2D keypoints}
\label{sec:method-fit}
The network's 2D keypoints are more accurate than the projection of its own 3D body: on 36k held-out frames of EgoHumans, the SimCC keypoints are off by 1.46\,\% of the crop side on average, and the projected 3D keypoints by 2.93\,\%.
An optional fit makes the 3D body agree with the better 2D.
It adjusts the root rotation, the 62 body angles and the camera translation, 68 unknowns grouped in $\mathbf{u}$, so that the 30 body keypoints of the rig, projected with the calibrated camera $\pi$, land on their SimCC estimates $\mathbf{x}_k$.
It does so by minimising the loss
\begin{equation}
  \mathcal{L}_{\mathrm{fit}}(\mathbf{u}) = \sum_{k=1}^{30} \frac{\big\lVert \pi\big(\mathbf{J}_k(\mathbf{u})\big) - \mathbf{x}_k \big\rVert^2}{\sigma_k^2} + \big\lVert \mathbf{u}-\mathbf{u}_0 \big\rVert^2_{\boldsymbol{\Lambda}}.
  \label{eq:fit}
\end{equation}
The first term is the reprojection error of each keypoint, divided by $\sigma_k$, the spread of its SimCC distributions, so that an uncertain keypoint counts less.
The second term is a prior: it keeps the solution close to the network's prediction $\mathbf{u}_0$, with a diagonal weight $\boldsymbol{\Lambda}$.
The body shape, the hands and the face are kept as predicted.
To minimise it, we take a single Levenberg-Marquardt step from $\mathbf{u}_0$.
Its Jacobian is exact: each angle rotates the joints below it about a known axis, so each derivative is a cross product.
ONNX has no operator to solve a linear system, and phone NPUs are slow on long chains of small operations, so the $68\times68$ system is solved by Newton-Schulz iterations, which use only matrix products.
The fit has no learned parameters, and it is exported in the same fp16 graph as the network: it runs entirely on the NPU and adds 0.9\,ms on the S23 and 0.14\,ms on the RTX 4070.
With two crops, the network also runs on a second, tighter crop of the same person, in the same batch.
Each 2D keypoint then has two estimates, one per crop, and we average them, giving more weight to the more confident one, \ie the one with the narrower SimCC distribution.
The fit starts from the body predicted on the first crop; the equations and the fp16 scaling are in the supplementary material, Sec.~\suppFit.

\paragraph{Crop scale.}
\method-M cuts a square crop with a side 1.2$\times$ the longer side of the person box.
For the fit variants, the box is first shrunk by a factor $s$, 0.9 for the first crop and 0.7 for the second, so that the person fills more of the crop.
Both factors were chosen on 3DPW validation and training data only (supplementary material, Sec.~\suppFit); EMDB-1 and RICH played no part, and the tighter crop improves them as well (\cref{tab:fit}).

%% file: tables/tab_speed.tex
\begin{tabular}{@{}l c rrrr rr rr}
\toprule
 & Accuracy (mm) & \multicolumn{8}{c}{Latency per person (ms)} \\
\cmidrule(lr){2-2}\cmidrule(l){3-10}
 & EMDB-1 & \multicolumn{3}{c}{RTX 4070} & CPU & \multicolumn{2}{c}{Phone NPU (fp16)} & \multicolumn{2}{c}{Phone CPU (fp32)} \\
\cmidrule(lr){3-5}\cmidrule(lr){6-6}\cmidrule(lr){7-8}\cmidrule(l){9-10}
Method & PA-MPJPE & TRT fp16 & ORT fp32 & PyTorch & i5 ORT & S23 & S24 & S23 & A14 \\
\midrule
\textcolor{ihmrblue}{\textbf{\method-M}} & \textcolor{ihmrblue}{48.8} & \textcolor{ihmrblue}{\textbf{0.98}} & \textcolor{ihmrblue}{4.42} & -- & \textcolor{ihmrblue}{\textbf{29.24}} & \cellcolor{ihmrzone}\textcolor{ihmrblue}{\textbf{5.49}} & \cellcolor{ihmrzone}\textcolor{ihmrblue}{\textbf{4.01}} & \textcolor{ihmrblue}{\textbf{92.97}} & \textcolor{ihmrblue}{\textbf{294}} \\
\textcolor{ihmrblue}{\textbf{\method-M + fit (1 crop)}} & \textcolor{ihmrblue}{45.2} & \textcolor{ihmrblue}{1.12} & \textcolor{ihmrblue}{5.78} & -- & \textcolor{ihmrblue}{30.74} & \cellcolor{ihmrzone}\textcolor{ihmrblue}{6.40} & \cellcolor{ihmrzone}\textcolor{ihmrblue}{4.67} & \textcolor{ihmrblue}{96.43} & \textcolor{ihmrblue}{300} \\
\textcolor{ihmrblue}{\textbf{\method-M + fit (2 crops)}} & \textcolor{ihmrblue}{44.4} & \textcolor{ihmrblue}{1.53} & \textcolor{ihmrblue}{6.85} & -- & \textcolor{ihmrblue}{56.35} & \cellcolor{ihmrzone}\textcolor{ihmrblue}{10.22} & \cellcolor{ihmrzone}\textcolor{ihmrblue}{7.00} & \textcolor{ihmrblue}{314} & \textcolor{ihmrblue}{568} \\
\addlinespace[2pt]\multicolumn{10}{@{}l}{\textcolor{ihmrgrey}{\footnotesize\textit{Run on a phone NPU}}} \\
SMPLer-X-S32 & 71.0 & 1.38\,{\scriptsize\textcolor{ihmrgrey}{1.4$\times$}} & 5.30 & 11.98 & 50.61 & \cellcolor{ihmrzone}11.12\,{\scriptsize\textcolor{ihmrgrey}{2$\times$}} & \cellcolor{ihmrzone}8.16 & 164 & 509 \\
MeshPose (HRNet-32) & 73.7 & 2.36\,{\scriptsize\textcolor{ihmrgrey}{2$\times$}} & 7.85 & 12.07 & 63.36 & 43.81\,{\scriptsize\textcolor{ihmrgrey}{8$\times$}} & \cellcolor{ihmrzone}31.19 & 309 & 979 \\
NLF-S & 45.0 & 1.52\,{\scriptsize\textcolor{ihmrgrey}{1.6$\times$}} & \textbf{3.71} & \textbf{10.19} & 33.16 & 67.97\,{\scriptsize\textcolor{ihmrgrey}{12$\times$}} & 54.45 & 103 & 306 \\
NLF-L & \underline{39.3} & 4.32\,{\scriptsize\textcolor{ihmrgrey}{4$\times$}} & 10.62 & 16.40 & 136 & 125\,{\scriptsize\textcolor{ihmrgrey}{23$\times$}} & 97.54 & 638 & 2,013 \\
Multi-HMR-S & 68.5 & 3.94\,{\scriptsize\textcolor{ihmrgrey}{4$\times$}} & 24.66 & 32.12 & 387 & 139\,{\scriptsize\textcolor{ihmrgrey}{25$\times$}} & 108 & 1,670 & 4,420 \\
\addlinespace[2pt]\multicolumn{10}{@{}l}{\textcolor{ihmrgrey}{\footnotesize\textit{Do not run on a phone NPU}}} \\
HybrIK-R34 & 69.2 & -- & -- & 14.95 & -- & -- & -- & -- & -- \\
CameraHMR & 46.7 & 5.31\,{\scriptsize\textcolor{ihmrgrey}{5$\times$}} & 13.87 & 21.61 & 408 & $\times$ & $\times$ & -- & $\times$ \\
TokenHMR & 54.7 & 5.31\,{\scriptsize\textcolor{ihmrgrey}{5$\times$}} & 14.56 & 23.36 & 409 & $\times$ & $\times$ & -- & $\times$ \\
HMR2.0b & 78.4 & 5.51\,{\scriptsize\textcolor{ihmrgrey}{6$\times$}} & 14.91 & 24.76 & 412 & $\times$ & $\times$ & -- & $\times$ \\
Fast SAM 3D Body & \textbf{38.6} & 130\,{\scriptsize\textcolor{ihmrgrey}{133$\times$}} & -- & 192 & -- & $\times$ & $\times$ & -- & $\times$ \\
SAM 3D Body & 40.3 & -- & -- & 416 & -- & $\times$ & $\times$ & -- & $\times$ \\
\bottomrule
\end{tabular}

%% file: tables/tab_accuracy.tex
\begin{tabular}{@{}l ccc ccc ccc rr}
\toprule
 & \multicolumn{3}{c}{EMDB-1 (J24, mm)} & \multicolumn{3}{c}{3DPW (J14, mm)} & \multicolumn{3}{c}{RICH (J24, mm)} & \multicolumn{2}{c}{Latency (ms)} \\
\cmidrule(lr){2-4}\cmidrule(lr){5-7}\cmidrule(lr){8-10}\cmidrule(l){11-12}
Method & PA & MPJPE & PVE & PA & MPJPE & PVE & PA & MPJPE & PVE & RTX 4070 & S23 NPU \\
\midrule
\textcolor{ihmrblue}{\textbf{\method-M}} & \textcolor{ihmrblue}{48.8} & \textcolor{ihmrblue}{74.9} & \textcolor{ihmrblue}{89.1} & \textcolor{ihmrblue}{40.0} & \textcolor{ihmrblue}{64.0} & \textcolor{ihmrblue}{76.8} & \textcolor{ihmrblue}{41.7} & \textcolor{ihmrblue}{64.7} & \textcolor{ihmrblue}{74.1} & \textcolor{ihmrblue}{\textbf{0.98}} & \cellcolor{ihmrzone}\textcolor{ihmrblue}{\textbf{5.49}} \\
\textcolor{ihmrblue}{\textbf{\method-M + fit (1 crop)}} & \textcolor{ihmrblue}{45.2} & \textcolor{ihmrblue}{70.9} & \textcolor{ihmrblue}{84.3} & \textcolor{ihmrblue}{39.4} & \textcolor{ihmrblue}{63.2} & \textcolor{ihmrblue}{75.1} & \textcolor{ihmrblue}{40.6} & \textcolor{ihmrblue}{63.0} & \textcolor{ihmrblue}{72.5} & \textcolor{ihmrblue}{1.12} & \cellcolor{ihmrzone}\textcolor{ihmrblue}{6.40} \\
\textcolor{ihmrblue}{\textbf{\method-M + fit (2 crops)}} & \textcolor{ihmrblue}{44.4} & \textcolor{ihmrblue}{69.8} & \textcolor{ihmrblue}{83.2} & \textcolor{ihmrblue}{39.3} & \textcolor{ihmrblue}{62.9} & \textcolor{ihmrblue}{74.8} & \textcolor{ihmrblue}{41.0} & \textcolor{ihmrblue}{64.4} & \textcolor{ihmrblue}{73.8} & \textcolor{ihmrblue}{1.53} & \cellcolor{ihmrzone}\textcolor{ihmrblue}{10.22} \\
\addlinespace[2pt]\multicolumn{12}{@{}l}{\textcolor{ihmrgrey}{\footnotesize\textit{Run on a phone NPU}}} \\
SMPLer-X-S32$^\dagger$$^\ddagger$ & 71.0 & 109.8 & 127.7 & 50.4 & 78.4 & 91.9 & 61.2 & 99.5 & 112.2 & 1.38\,{\scriptsize\textcolor{ihmrgrey}{1.4$\times$}} & \cellcolor{ihmrzone}11.12\,{\scriptsize\textcolor{ihmrgrey}{2$\times$}} \\
MeshPose (HRNet-32)$^\dagger$ & 73.7 & -- & -- & 49.7 & -- & -- & -- & -- & -- & 2.36\,{\scriptsize\textcolor{ihmrgrey}{2$\times$}} & 43.81\,{\scriptsize\textcolor{ihmrgrey}{8$\times$}} \\
NLF-S$^\ddagger$ & 45.0 & 74.6 & 87.9 & 39.4 & 62.2 & 74.6 & 33.7 & 56.9 & 65.7 & 1.52\,{\scriptsize\textcolor{ihmrgrey}{1.6$\times$}} & 67.97\,{\scriptsize\textcolor{ihmrgrey}{12$\times$}} \\
NLF-L$^\ddagger$ & \underline{39.3} & \underline{67.7} & \underline{79.7} & \underline{35.1} & \underline{57.2} & \underline{68.3} & \textbf{27.3} & \textbf{47.3} & \textbf{53.7} & 4.32\,{\scriptsize\textcolor{ihmrgrey}{4$\times$}} & 125\,{\scriptsize\textcolor{ihmrgrey}{23$\times$}} \\
Multi-HMR-S & 68.5 & 103.0 & 115.6 & 54.8 & 84.4 & 99.1 & 69.4 & 102.4 & 112.5 & 3.94\,{\scriptsize\textcolor{ihmrgrey}{4$\times$}} & 139\,{\scriptsize\textcolor{ihmrgrey}{25$\times$}} \\
\addlinespace[2pt]\multicolumn{12}{@{}l}{\textcolor{ihmrgrey}{\footnotesize\textit{Do not run on a phone NPU}}} \\
HybrIK-R34$^\dagger$ & 69.2 & 104.7 & 126.7 & 49.0 & 76.2 & 92.3 & 59.4 & 100.2 & 115.1 & 14.95$^p$\,{\scriptsize\textcolor{ihmrgrey}{15$\times$}} & -- \\
CameraHMR & 46.7 & 76.7 & 91.5 & 38.8 & 61.3 & 72.6 & 33.6 & 54.8 & 63.8 & 5.31\,{\scriptsize\textcolor{ihmrgrey}{5$\times$}} & $\times$ \\
TokenHMR & 54.7 & 88.8 & 106.5 & 44.5 & 69.8 & 83.5 & 42.1 & 80.6 & 93.6 & 5.31\,{\scriptsize\textcolor{ihmrgrey}{5$\times$}} & $\times$ \\
HMR2.0b & 78.4 & 117.3 & 140.1 & 54.1 & 80.8 & 92.1 & 58.0 & 104.8 & 122.3 & 5.51\,{\scriptsize\textcolor{ihmrgrey}{6$\times$}} & $\times$ \\
Fast SAM 3D Body$^\dagger$ & \textbf{38.6} & 65.5 & 77.4 & \textbf{34.1} & \textbf{55.0} & \textbf{66.3} & 32.1 & 53.1 & 60.5 & 130$^t$\,{\scriptsize\textcolor{ihmrgrey}{133$\times$}} & $\times$ \\
SAM 3D Body$^\dagger$ & 40.3 & \textbf{65.1} & \textbf{77.4} & 35.1 & 55.7 & 67.2 & 33.5 & 54.6 & 62.3 & 416$^p$\,{\scriptsize\textcolor{ihmrgrey}{425$\times$}} & $\times$ \\
\bottomrule
\end{tabular}

%% file: sec/4_protocol.tex
\section{Evaluation protocol}
\label{sec:protocol}

\begin{table}[t]
  \centering
  \small
  \setlength{\tabcolsep}{3pt}
  \resizebox{\columnwidth}{!}{\input{tables/tab_family}}
  \caption{\textbf{The \method family.} Nano, tiny, small and medium use HGNetV2-B0, B1, B2 and B4 backbones; \emph{M-C} is the medium architecture trained only on commercially usable images (Harmony4D and COCO). Protocol of \cref{tab:accuracy}. Latency per person: RTX~4070 with TensorRT fp16, Samsung Galaxy S23 NPU in fp16 and CPU in fp32. \colorbox{ihmrzone}{Shaded}: real time on the phone ($\leq$\,33.3\,ms). \textbf{+ fit}: \cref{sec:method-fit}, on one or two crops; medium's two-crop graph is disproportionately slow on the S23 CPU (three jobs agree).}
  \label{tab:family}
\end{table}

\begin{figure*}[t]
  \centering
  \includegraphics[width=\textwidth]{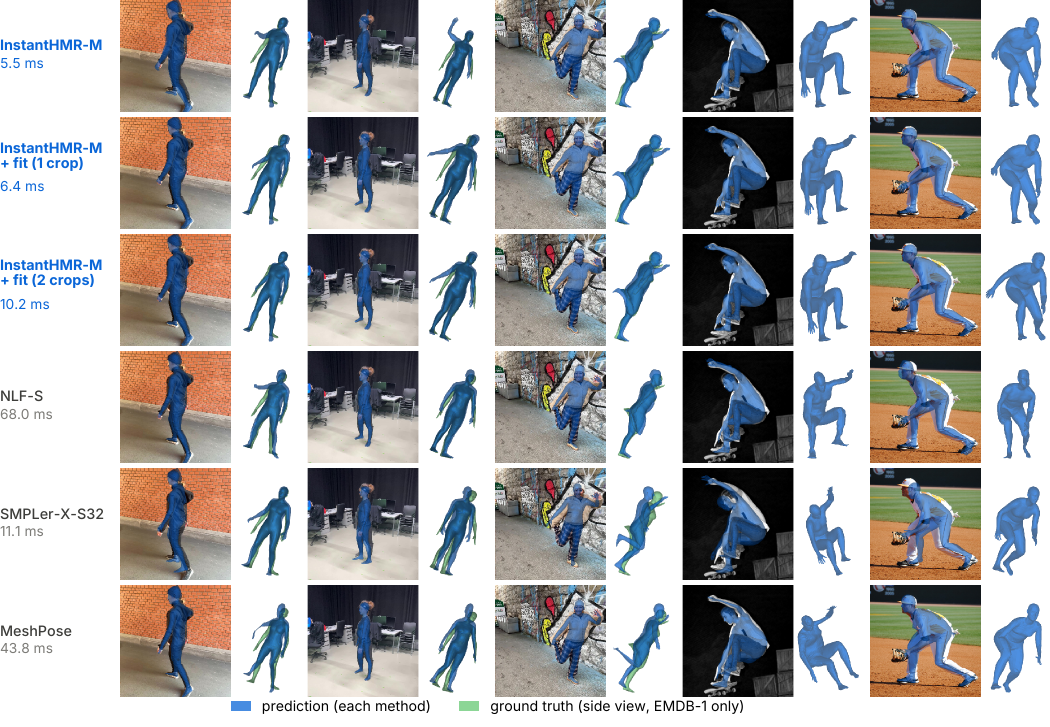}
  \caption{\textbf{Qualitative comparison of the methods that run on a phone NPU.} For each image: the mesh projected into the image, and the same mesh seen from the side. On EMDB-1~\cite{kaufmann2023emdb} (columns 1--3), the side view also shows the ground truth in green, with the prediction placed on its hips and scaled to its size, but not rotated; the COCO~\cite{lin2014coco} images (columns 4--5) have no 3D ground truth. Under each method: its latency on the S23 NPU. Rows 2--3: \method-M with the kinematic fit on one and two crops, which lowers its PA-MPJPE from 91.5 to 49.3 and 35.2\,mm in column 1, and from 182.2 to 87.1 and 57.3\,mm in column 2.}
  \label{fig:qualitative}
\end{figure*}

We fix one protocol, the one used by the MHR-based SAM~3D~Body and Fast SAM~3D~Body, and score every method under it.

\paragraph{Settings.}
Every method is scored on the same frames, with the same person boxes and the same camera.
The person boxes are the ones the benchmarks provide: EMDB ships one per frame, and for 3DPW, which ships none, we take the extent of the annotated 2D keypoints enlarged by 20\,\%.
Each method cuts its own crop from these boxes, with the preprocessing its authors released; ours, including the crop scale of the fit variants, was fixed on validation data before we scored the test sets (\cref{sec:method-fit}).
The only exception is NLF, which we run with its own person detector, as in its paper; giving it the ground-truth boxes instead changes its error by less than 1\,mm.
The focal length is the calibrated one of each benchmark.
Methods that output SMPL are scored as they are, and the others are first converted to SMPL: MHR outputs with Meta's official MHR-to-SMPL fitting, as in the Fast SAM~3D~Body paper, and SMPL-X outputs through Meta's surface maps from SMPL-X to MHR to SMPL, which add 0.67\,mm of error on average.
As in each benchmark's literature, EMDB-1 is scored on the 24 SMPL joints and 3DPW on the 14 joints regressed from the mesh with the Human3.6M regressor.
For the three methods that do not output SMPL (SMPLer-X, Multi-HMR and MeshPose), the 24 EMDB-1 joints are regressed from the vertices.
We score every 10th frame, which gives 2,419 frames on EMDB-1 and 3,562 person-frames on 3DPW; scoring every frame instead changes PA-MPJPE by at most 0.3\,mm.
On RICH~\cite{huang2022rich}, we score the 21,287 test crops of CameraHMR's label file~\cite{patel2025camerahmr}, with its person boxes, against the SMPL-X ground truth converted to SMPL (24 joints).
A frame on which a method returns no prediction counts as missing.

\paragraph{Reproduction.}
We run each competitor from its released code and weights, with its own crop and preprocessing, and call it \emph{reproduced} when its PA-MPJPE is within 1\,mm of its paper's, on that paper's joint set.
NLF-S, TokenHMR and HMR2.0b reproduce on both EMDB-1 and 3DPW, and CameraHMR on 3DPW and RICH; the other gaps to the papers range from 2.2\,mm better to 4.5\,mm worse on EMDB-1 and 3DPW, and up to 9.9\,mm worse on RICH (supplementary material, Sec.~\suppReproduction). SMPLer-X-S32 and Multi-HMR-S have no published number for their released variant.
Such gaps can come from settings that a paper does not fully specify, and we do not claim that any published number is wrong.

\paragraph{Latency.}
Latency is the model's time per person crop at batch 1, from the crop to a posed mesh of up to 6,890 vertices; detection and cropping are excluded for every method.
Desktop timings are medians over interleaved rounds, so background load affects all methods alike.
Phone timings use Qualcomm AI Hub~\cite{qualcommaihub} on physical devices, the median of three jobs; they cover the model alone, and \cref{sec:experiments} adds end-to-end timings in an Android app.
Only complete methods are timed: a method that does not export whole is timed in its own release (HybrIK, SAM~3D~Body, Fast SAM~3D~Body), and partial timings are left to the supplementary material, Sec.~\suppPartial.

%% file: tables/tab_family.tex
\begin{tabular}{@{}l ccc c cc@{}}
\toprule
 & \multicolumn{3}{c}{PA-MPJPE (mm)} & \multicolumn{3}{c}{Latency (ms)} \\
\cmidrule(lr){2-4}\cmidrule(l){5-7}
 & & & & RTX & \multicolumn{2}{c}{S23} \\
\cmidrule(l){6-7}
Model & EMDB-1 & 3DPW & RICH & 4070 & NPU & CPU \\
\midrule
\method-N & 61.7 & 46.6 & 53.5 & 0.70 & \cellcolor{ihmrzone}4.60 & \cellcolor{ihmrzone}31.5 \\
\quad + fit, 1 crop & 57.2 & 44.5 & 49.9 & 0.85 & \cellcolor{ihmrzone}5.49 & 36.0 \\
\quad + fit, 2 crops & 55.9 & 44.1 & 49.5 & 1.10 & \cellcolor{ihmrzone}8.72 & 81.9 \\
\method-T & 58.0 & 44.0 & 49.7 & 0.73 & \cellcolor{ihmrzone}4.84 & 37.4 \\
\quad + fit, 1 crop & 53.8 & 42.4 & 47.1 & 0.87 & \cellcolor{ihmrzone}5.81 & 41.4 \\
\quad + fit, 2 crops & 52.6 & 42.0 & 47.1 & 1.13 & \cellcolor{ihmrzone}9.22 & 96.6 \\
\method-S & 53.3 & 41.7 & 45.4 & 0.84 & \cellcolor{ihmrzone}5.38 & 54.5 \\
\quad + fit, 1 crop & 49.2 & 40.3 & 43.5 & 1.00 & \cellcolor{ihmrzone}6.37 & 57.3 \\
\quad + fit, 2 crops & 48.1 & 40.0 & 43.7 & 1.30 & \cellcolor{ihmrzone}10.37 & 154.2 \\
\textcolor{ihmrblue}{\textbf{\method-M}} & \textcolor{ihmrblue}{48.8} & \textcolor{ihmrblue}{40.0} & \textcolor{ihmrblue}{41.7} & \textcolor{ihmrblue}{0.98} & \cellcolor{ihmrzone}\textcolor{ihmrblue}{5.49} & \textcolor{ihmrblue}{93.0} \\
\textcolor{ihmrblue}{\quad + fit, 1 crop} & \textcolor{ihmrblue}{45.2} & \textcolor{ihmrblue}{39.4} & \textcolor{ihmrblue}{40.6} & \textcolor{ihmrblue}{1.12} & \cellcolor{ihmrzone}\textcolor{ihmrblue}{6.40} & \textcolor{ihmrblue}{96.4} \\
\textcolor{ihmrblue}{\quad + fit, 2 crops} & \textcolor{ihmrblue}{44.4} & \textcolor{ihmrblue}{39.3} & \textcolor{ihmrblue}{41.0} & \textcolor{ihmrblue}{1.53} & \cellcolor{ihmrzone}\textcolor{ihmrblue}{10.22} & \textcolor{ihmrblue}{314.0} \\
\midrule
\method-M-C & 72.2 & 50.7 & 65.5 & 0.98 & \cellcolor{ihmrzone}5.49 & 93.0 \\
\quad + fit, 1 crop & 68.8 & 47.7 & 63.1 & 1.12 & \cellcolor{ihmrzone}6.40 & 96.4 \\
\quad + fit, 2 crops & 68.3 & 47.5 & 64.7 & 1.53 & \cellcolor{ihmrzone}10.22 & 314.0 \\
\bottomrule
\end{tabular}

%% file: sec/5_experiments.tex
\section{Experiments}
\label{sec:experiments}

We evaluate on EMDB-1~\cite{kaufmann2023emdb}, 3DPW test~\cite{vonmarcard2018pw3d} and RICH test~\cite{huang2022rich} under the protocol of \cref{sec:protocol}. None of the three benchmarks' images are in \method's training data; five competitors trained on 3DPW-train (marked $^\dagger$), and three on RICH ($^\ddagger$), which makes EMDB-1 the fairer of the three.

\paragraph{Speed (\cref{tab:speed}).}
On the Galaxy S23 NPU, \method-M takes 5.49\,ms per person (4.01\,ms on the S24), and 6.40 and 10.22\,ms with the one- and two-crop fit; all three are real time.
The only other method that is real time on the phone, SMPLer-X-S32 (11.12\,ms), is 22\,mm less accurate on EMDB-1, and every ViT-H or DINOv3-based method (CameraHMR, TokenHMR, HMR2.0, SAM~3D~Body, Fast SAM~3D~Body) fails to run on either phone NPU, for lack of memory or because the compiled model fails.
On the desktop GPU, \method-M is the fastest method with TensorRT (0.98\,ms; nano: 0.70\,ms), and with the one-crop fit (1.12\,ms) it is still faster than every other method.

\paragraph{In an Android app.}
We run \method-M with the one-crop fit in an Android app on the S23 (ONNX Runtime with its QNN back end, fp16, every layer on the NPU), tracking one person, with a NanoDet-Plus person detector~\cite{nanodet} run on the NPU once per second.
Replaying a 720$\times$960 EMDB video at 30\,FPS on battery, a frame takes 12.73\,ms (median; 13.46\,ms at the 95th percentile): 5.21\,ms for the crop on the CPU and 7.41\,ms for the model with the fit, the 3D keypoints and a 595-vertex mesh, 1.0\,ms more than on AI Hub, mostly for tensor copies; frames with a detection take 31.4\,ms.
The app kept 30.0\,FPS for five minutes (2 of 8,996 frames dropped) and over 342\,s of live camera (1 of 10,409 dropped).
Its fp16 output stays within 2.9\,mm (mean) of the fp32 graph in 3D, so the accuracy of \cref{tab:accuracy} holds on the phone; power and per-stage details are in the supplementary material, Sec.~\suppDevice.

\paragraph{Why \method is fast on the NPU.}
On the S23 CPU, \method-M (93\,ms) and NLF-S (103\,ms) are close, but on the NPU NLF-S is 12$\times$ slower, although both run entirely on it.
NLF estimates each of the 6,890 vertices as a separate query point with its own $8\times8\times8$ heatmap, so most of its work is moving and summing $6{,}890\times512$ tensors, which an NPU, built for convolutions and matrix products, does slowly: gathers, reductions, transposes, reshapes and softmaxes take 71\% of its NPU time, its convolutions under 2\%.
\method regresses the 204 MHR parameters once per person, and the mesh follows from the rig's forward kinematics and skinning, a fixed weighted sum.

\paragraph{Accuracy (\cref{tab:accuracy}).}
Among the methods that run on a phone, \method-M is 20--25\,mm more accurate than SMPLer-X, MeshPose and Multi-HMR on EMDB-1 (PA-MPJPE), 10--15\,mm on 3DPW and 20--28\,mm on RICH, and the gap can be much larger on hard poses (\cref{fig:qualitative}).
MeshPose is released only with its HRNet-32 backbone; its paper reports its unreleased phone-sized variants 2--3\,mm behind it on 3DPW\footnote{MeshPose-S / -XS, as published~\cite{le2024meshpose}: 48.97 / 49.71\,mm 3DPW PA-MPJPE on 14 LSP joints; not comparable with our rows.}.
Only NLF is more accurate, at 12$\times$ (NLF-S) and 23$\times$ (NLF-L) the phone latency.
The kinematic fit closes most of this gap.
With one crop, \method-M is within 0.2\,mm of NLF-S's PA-MPJPE on EMDB-1 (45.2 against 45.0\,mm), with a lower MPJPE and PVE, and equal to it on 3DPW (39.4\,mm), 10.6$\times$ faster on the phone; with two crops, it beats NLF-S on all three EMDB-1 metrics, 6.7$\times$ faster.
On EMDB-1 and 3DPW, the only method that runs on a phone NPU and is clearly more accurate is then NLF-L, at 125\,ms.
On RICH, on which NLF is trained, the fit improves \method-M from 41.7 to 40.6\,mm, still behind NLF.
Without the phone constraint, Fast SAM~3D~Body is the most accurate on EMDB-1 and 3DPW, and SAM~3D~Body is close to NLF-L; both are 8--10\,mm better than \method-M on EMDB-1, at 130--416\,ms per person on a desktop GPU.

\paragraph{The family (\cref{tab:family}).}
From nano to medium, EMDB-1 PA-MPJPE improves from 61.7 to 48.8\,mm, while the S23 NPU latency only grows from 4.60 to 5.49\,ms, so medium is the natural choice on the NPU.
On the S23 CPU, the latency triples (31.5 to 93.0\,ms) and only nano runs in real time, so the smaller sizes suit phones without a usable NPU.
The fit helps every size, the smaller ones most: with two crops, it gains 5.8\,mm (nano) to 4.4\,mm (medium) on EMDB-1, 2.5 to 0.7\,mm on 3DPW and 4.0 to 0.7\,mm on RICH.
Training on commercially usable images only (M-C: Harmony4D and COCO, 0.5\,M of the 4.2\,M crops) costs 23\,mm on EMDB-1: the training images matter more than any architectural change we tried.

\paragraph{Crop scale and kinematic fit (\cref{tab:fit}).}
The tighter crop and the fit are separate effects whose gains roughly add up: on EMDB-1, the crop alone gains 2.2\,mm, the fit alone 1.3\,mm, and both 3.6\,mm.
On top of the tighter crop, the fit gains 1.4\,mm on EMDB-1 and 0.6\,mm on RICH but is neutral on 3DPW (0.2\,mm worse), and the second crop gains another 0.8\,mm on EMDB-1 but loses 0.4\,mm on RICH.
The fit is limited by the 2D keypoints: on two EgoHumans sequences, the fit gains 0.4 and 0.7\,mm with the network's own 2D keypoints, but would gain 17.1 and 6.1\,mm with the ground-truth ones (supplementary material, Sec.~\suppFit).

\begin{table}[t]
  \centering
  \small
  \input{tables/tab_fit_ablation}
  \caption{\textbf{Crop scale and kinematic fit.} PA-MPJPE (mm) under the protocol of \cref{tab:accuracy} (EMDB-1 and RICH on 24 joints, 3DPW on 14), from the exported TensorRT fp16 graphs. Crop scale: the factor applied to the person box before the 1.2$\times$ square crop (\cref{sec:method-fit}). The first row is \method-M as in \cref{tab:accuracy}; the last two are its two fit variants.}
  \label{tab:fit}
\end{table}

\paragraph{Temporal stability.}
\method estimates each frame on its own, so its output jitters on video.
Over every frame of EMDB-1, \method-M has an acceleration error of 27.1\,mm/frame$^2$, against 21.2 for NLF-S.
The two-crop fit lowers it to 22.5, and a causal 1 Euro filter~\cite{casiez2012oneeuro} on the fit's pose, tuned on validation data, to 15.2 (15.7 on 3DPW, against 18.4 for NLF-S), with no added latency and a 0.1--0.3\,mm better PA-MPJPE.
No filter is used in \cref{tab:accuracy,tab:fit}; Sec.~\suppTemporal{} of the supplementary material gives the details and a smoother setting.

%% file: tables/tab_fit_ablation.tex
\begin{tabular}{@{}l ccc@{}}
\toprule
Crop scale and fit & EMDB-1 & 3DPW & RICH \\
\midrule
1.0, no fit (\cref{tab:accuracy}) & 48.8 & 40.0 & 41.7 \\
0.9, no fit & 46.6 & \textbf{39.2} & 41.2 \\
1.0, fit & 47.5 & 40.0 & 41.1 \\
0.9, fit (1 crop) & 45.2 & 39.4 & \textbf{40.6} \\
0.9 and 0.7, fit (2 crops) & \textbf{44.4} & 39.3 & 41.0 \\
\bottomrule
\end{tabular}

%% file: sec/6_conclusion.tex
\section{Limitations and conclusion}
\label{sec:conclusion}

\paragraph{Limitations.}
\method has no dedicated hand branch: the hands are read from the same $288\times288$ crop and $9\times9$ image tokens as the rest of the body, where a hand covers only a few pixels, whereas SAM~3D~Body refines them with a second decoder on hand crops.
All our phone measurements are on Android phones with Qualcomm NPUs; we did not evaluate Apple's Neural Engine.
\method estimates each frame on its own, so smooth motion needs a causal filter, and the filters we tested trade smoothness against accuracy.
Finally, many training images, such as those of SA-1B and AI~Challenger, may only be used for research, so only \method-M-C can be used in commercial products, and it is 23\,mm less accurate on EMDB-1; closing this gap requires more permissively licensed training images, real or synthetic.
A light hand branch, a temporal model that adds no latency, and Apple silicon are left for future work.

\paragraph{Conclusion.}
\method-M is the most accurate method that runs in real time on a phone NPU, 12$\times$ faster than NLF-S, the faster of the two more accurate methods that also run there.
With a kinematic fit to its own 2D keypoints, exported in the same graph, it matches NLF-S on EMDB-1 and 3DPW at 10.6$\times$ its speed.
To make these comparisons fair, we re-evaluated eleven methods under one protocol and reported how closely each reproduces its paper.

%% file: sec/llm_statement.tex
\section*{LLM Usage Statement}
We used Claude Code (Anthropic), an LLM-based coding assistant, to write and run evaluation code under our direction (competitor reproduction, scoring, latency measurement, and the generation of tables), to search the literature, and to draft and edit text. The model, its training and the experimental design are ours; we reviewed all code, verified every source and revised all text, and every number comes from saved result files.

%% file: sec/supp_body.tex
\renewcommand{\thesection}{\Alph{section}}

\section{Network equations}
\label{sec:supp-network}
This section gives the equations behind Sec.~3.1 of the paper, with its notation: the crop $\mathbf{I}\in\mathbb{R}^{288\times288\times3}$, the crop position $\mathbf{c}\in\mathbb{R}^{3}$, the MHR rig parameters $\boldsymbol{\theta}\in\mathbb{R}^{204}$ and identity coefficients $\boldsymbol{\beta}\in\mathbb{R}^{45}$, the SimCC distributions $\mathbf{P}\in[0,1]^{70\times2\times96}$, the 2D keypoints $\mathbf{x}\in\mathbb{R}^{70\times2}$, the camera translation $\mathbf{t}\in\mathbb{R}^{3}$, the MHR rig $\mathcal{M}$, the posed mesh $\mathbf{V}$ and its 3D keypoints $\mathbf{J}\in\mathbb{R}^{70\times3}$.

\paragraph{Position in the full frame.}
For a person box of centre $(b_x, b_y)$ and size $b_w\times b_h$ in an image of size $W\times H$ taken with focal length $f$,
\begin{equation}
  \mathbf{c} = \Big(\arctan\frac{b_x - W/2}{f},\ \arctan\frac{b_y - H/2}{f},\ \frac{\max(b_w, b_h)}{f}\Big).
\end{equation}

\paragraph{Backbone and decoder.}
The backbone $E$ (HGNetV2, a linear projection to $d=512$ channels and a fixed 2D sine-cosine positional embedding) gives the image tokens $\mathbf{F}=E(\mathbf{I})\in\mathbb{R}^{81\times d}$.
A two-layer MLP $\phi$ embeds the crop position, and the embedding is added to each of the 71 learned queries $\mathbf{Q}\in\mathbb{R}^{71\times d}$.
The transformer decoder $D$ outputs one feature per query,
\begin{equation}
  \mathbf{Z} = D\big(\mathbf{Q} + \mathbf{1}\,\phi(\mathbf{c})^{\top},\ \mathbf{F}\big) \in \mathbb{R}^{71\times d},
\end{equation}
where $\mathbf{1}\in\mathbb{R}^{71}$ is a vector of ones, $\mathbf{z}_0$ is the row of the body query and $\mathbf{z}_1,\dots,\mathbf{z}_{70}$ are the rows of the keypoint queries.

\paragraph{Heads.}
\begin{equation}
  (\boldsymbol{\theta},\boldsymbol{\beta}) = h_{\mathrm{MHR}}(\mathbf{z}_0), \qquad
  \mathbf{t} = h_{\mathrm{cam}}(\mathbf{z}_0), \qquad
  \mathbf{P}_k = h_{\mathrm{2D}}(\mathbf{z}_k), \quad k=1,\dots,70.
\end{equation}
With $u_1,\dots,u_{96}$ the bin centres, spread uniformly over $[-1.5, 1.5]$ in normalised crop units ($[-1,1]$ across the crop), the 2D keypoints are the expectations
\begin{equation}
  \mathbf{x}_{k} = \Big(\sum_{b=1}^{96} P_{k,x,b}\,u_b,\ \sum_{b=1}^{96} P_{k,y,b}\,u_b\Big).
\end{equation}

\paragraph{MHR head.}
The MHR head $h_{\mathrm{MHR}}$ is a two-layer head in SAM~3D~Body's continuous parameterisation: the root rotation as a 6D vector, the body pose in its continuous form, the identity coefficients $\boldsymbol{\beta}$, 28 bone-scale principal components and two hand blocks, which are decoded into the rig parameters $\boldsymbol{\theta}$.
The root translation is fixed to zero, since the camera translation places the person.
The bone-scale coefficients and the six limb-length parameters are bounded to the range seen in training.
Without these bounds, the scales enter the rig's forward kinematics multiplicatively along each limb, so a single large prediction can produce a skeleton hundreds of metres tall and make the training diverge.

\paragraph{From the crop to the posed mesh.}
Putting it together, the network $\mathcal{N}$ maps the crop and its position to the posed mesh in camera coordinates,
\begin{equation}
  \mathcal{N}(\mathbf{I}, \mathbf{c}) = \mathcal{M}\big(h_{\mathrm{MHR}}(\mathbf{z}_0)\big) + \mathbf{1}\,h_{\mathrm{cam}}(\mathbf{z}_0)^{\top},
\end{equation}
where the translation is added to every vertex.

\section{Losses}
\label{sec:supp-losses}
We write $\cdot^{\star}$ for an annotation, $\mathbf{R}$ for the root rotation matrix, $\mathbf{s}$ for the 68 bone scales in $\boldsymbol{\theta}$, and $\boldsymbol{\theta}_{\mathrm{loc}}$ for its 130 joint parameters.
The loss is $\mathcal{L}=\sum_i \lambda_i \mathcal{L}_i$ over the eleven terms of \cref{tab:supp-losses}.
The 3D terms compare the 70 keypoints and a farthest-point subset $\mathcal{S}$ of 595 vertices of the full MHR mesh with unsquared Euclidean distances:
\begin{equation}
  \mathcal{L}_{\mathbf{J}} = \frac{\sum_k w_k \lVert \mathbf{J}_k - \mathbf{J}^{\star}_k \rVert_2}{\sum_k w_k}, \qquad
  \mathcal{L}_{\mathbf{V}} = \frac{1}{|\mathcal{S}|}\sum_{v\in\mathcal{S}} \lVert \mathbf{V}_v - \mathbf{V}^{\star}_v \rVert_2,
  \label{eq:loss3d}
\end{equation}
where $w_k=0.2$ for the 40 finger keypoints and 1 otherwise.
The SimCC head is trained with a cross-entropy against a Gaussian label $\mathbf{q}$ centred on the annotated coordinate, with a standard deviation of 2 bins,
\begin{equation}
  \mathcal{L}_{\mathrm{CE}} = -\frac{1}{140}\sum_{k=1}^{70}\sum_{a\in\{x,y\}}\sum_{b=1}^{96} q_{k,a,b}\,\log P_{k,a,b},
  \label{eq:lossce}
\end{equation}
and the reprojection term ties the 3D keypoints to the image through the calibrated camera projection $\pi$, expressed in normalised crop units:
\begin{equation}
  \mathcal{L}_{\pi} = \frac{\sum_k w_k \lVert \pi(\mathbf{J}_k + \mathbf{t}) - \mathbf{x}^{\star}_k \rVert_1}{\sum_k w_k}.
  \label{eq:lossreproj}
\end{equation}
The terms that go through the rig ($\mathcal{L}_{\mathbf{J}}$, $\mathcal{L}_{\mathbf{V}}$, $\mathcal{L}_{\pi}$) are multiplied by a factor that grows linearly from 0 to 1 over the first 2,000 steps.
The parameter terms skip horizontally flipped samples, whose annotated parameters are not mirrored, and the root rotation also skips rotated samples.
The weights were tuned empirically, with single-batch overfitting tests and by balancing the measured gradient norm of each term.

\begin{table}[h]
  \centering
  \begin{tabular}{@{}llr@{}}
    \toprule
    Term & Form & $\lambda$ \\
    \midrule
    $\mathcal{L}_{\boldsymbol{\theta}}$, joint parameters $\boldsymbol{\theta}_{\mathrm{loc}}$ & Smooth-L1 ($\beta$\,=\,0.05\,rad) & 0.96 \\
    $\mathcal{L}_{\mathbf{R}}$, root rotation & $\tfrac14\lVert\mathbf{R}-\mathbf{R}^{\star}\rVert_F^2$ & 0.0074 \\
    $\mathcal{L}_{\mathbf{s}}$, bone scales & MSE & 0.1 \\
    $\mathcal{L}_{\boldsymbol{\beta}}$, identity coefficients & MSE & 0.03 \\
    $\mathcal{L}_{\mathbf{t}}$, camera translation & MSE & 0.1 \\
    $\mathcal{L}_{\mathbf{J}}$, 70 3D keypoints & Euclidean, \cref{eq:loss3d} & 0.35 \\
    $\mathcal{L}_{\mathbf{V}}$, 595 vertices & Euclidean, \cref{eq:loss3d} & 0.35 \\
    $\mathcal{L}_{\mathrm{skel}}$, 127 skeleton joints & Smooth-L1 & 0.001 \\
    $\mathcal{L}_{\mathrm{CE}}$, SimCC & cross-entropy, \cref{eq:lossce} & 1 \\
    $\mathcal{L}_{\mathbf{x}}$, 2D keypoints $\mathbf{x}$ & Smooth-L1 & 10 \\
    $\mathcal{L}_{\pi}$, reprojection & L1, \cref{eq:lossreproj} & 0.01 \\
    \bottomrule
  \end{tabular}
  \caption{\textbf{Training losses and their weights.} 3D distances in metres, 2D coordinates in normalised crop units ($[-1,1]$ across the crop).}
  \label{tab:supp-losses}
\end{table}

\section{How much the evaluation setting moves one model}
With every other setting fixed, the way MHR is converted to SMPL changes \method-M's EMDB-1 PA-MPJPE by up to 1.1\,mm across the three routes that fit or map the mesh: 48.8\,mm with Meta's official fitter, 49.1 with our own SMPL fit and 49.9 with a fixed surface map. Fast SAM~3D~Body's learned conversion network is much worse, at 65.8\,mm.
The focal length barely matters for PA-MPJPE (48.9\,mm with a fixed 55$^\circ$ field of view), and detector boxes help by 1.9\,mm, because they are tighter (\cref{sec:supp-det}).
These spreads are as large as the gaps between methods in the paper's Tab.~2, which is why every row there shares one protocol (\cref{tab:supp-protocol}; the same settings for the other sizes are in \cref{sec:supp-protocol-models}).

\begin{table}[h]
  \centering
  \input{tables/tab_protocol_ablation}
  \caption{\textbf{How much the evaluation setting moves one model.} \method-M under every combination of person boxes (ground truth, matched detections, or automatic detections where a missed person counts as missing), focal length (calibrated, or a fixed 55$^\circ$ field of view) and MHR-to-SMPL route. First row, in blue: the protocol of the paper's Tab.~2. Errors in mm.}
  \label{tab:supp-protocol}
\end{table}

\section{Protocol settings for every size}
\label{sec:supp-protocol-models}
\Cref{tab:supp-protocol-models} repeats the main settings of \cref{tab:supp-protocol} for every size of \method.
Each column changes one setting of the paper's protocol and keeps the others.
Detector boxes improve every size, by 1.7--2.3\,mm on EMDB-1 and 0.7--1.4\,mm on 3DPW, and \method-M-C by 1.9 and 2.9\,mm (\cref{sec:supp-det} explains why).
A fixed field of view changes PA-MPJPE by at most 0.4\,mm.
The two other routes from MHR to SMPL make every size worse: the surface map alone by 0.9--1.1\,mm on EMDB-1 and 1.0--1.4\,mm on 3DPW, and Fast SAM~3D~Body's learned conversion network by 11.7--17.0\,mm and 2.5--3.8\,mm.
Every size therefore reacts to the evaluation setting in the same way as \method-M.

\begin{table}[h]
  \centering
  \small
  \resizebox{\linewidth}{!}{%
  \begin{tabular}{@{}l cc cc cc cc cc@{}}
    \toprule
    & \multicolumn{2}{c}{Paper's protocol} & \multicolumn{2}{c}{Detector boxes} & \multicolumn{2}{c}{55$^\circ$ field of view} & \multicolumn{2}{c}{Surface map} & \multicolumn{2}{c}{Fast's network} \\
    \cmidrule(lr){2-3}\cmidrule(lr){4-5}\cmidrule(lr){6-7}\cmidrule(lr){8-9}\cmidrule(l){10-11}
    Model & EMDB-1 & 3DPW & EMDB-1 & 3DPW & EMDB-1 & 3DPW & EMDB-1 & 3DPW & EMDB-1 & 3DPW \\
    \midrule
    \method-N & 61.7 & 46.6 & 59.4 & 45.2 & 61.6 & 46.6 & 62.6 & 47.6 & 73.4 & 49.1 \\
    \method-T & 58.0 & 44.0 & 56.3 & 43.0 & 57.7 & 44.1 & 59.1 & 45.2 & 71.0 & 46.8 \\
    \method-S & 53.3 & 41.7 & 51.4 & 40.3 & 52.9 & 41.7 & 54.3 & 42.9 & 67.9 & 45.0 \\
    \textcolor{ihmrblue}{\textbf{\method-M}} & \textcolor{ihmrblue}{48.8} & \textcolor{ihmrblue}{40.0} & 46.9 & 39.3 & 48.9 & 40.0 & 49.9 & 41.4 & 65.8 & 43.8 \\
    \midrule
    \method-M-C & 72.2 & 50.7 & 70.3 & 47.8 & -- & -- & -- & -- & -- & -- \\
    \bottomrule
  \end{tabular}}
  \caption{\textbf{Protocol settings for every size.} PA-MPJPE in mm, EMDB-1 on 24 joints and 3DPW on 14. The paper's protocol uses the ground-truth boxes, the calibrated focal length and Meta's official MHR-to-SMPL fitting. Each other column changes one of these: the matched detector boxes of \cref{tab:supp-family}; a fixed 55$^\circ$ field of view; the MHR-to-SMPL surface map without fitting; Fast SAM~3D~Body's learned conversion network. \method-M-C was scored in the first two settings only.}
  \label{tab:supp-protocol-models}
\end{table}

\section{Every frame against every 10th frame}
\label{sec:supp-stride}
The paper scores every 10th frame of EMDB-1 and 3DPW (Sec.~4).
\Cref{tab:supp-stride} also scores \method-M and NLF-S on every frame.
PA-MPJPE moves by at most 0.30\,mm (\method-M on 3DPW, 24 joints), and by at most 0.06\,mm on EMDB-1.
On each benchmark, both methods move in the same direction, so the subsampling is small against the gaps between methods.

\begin{table}[h]
  \centering
  \small
  \resizebox{\linewidth}{!}{\input{tables/supp_stride_check}}
  \caption{\textbf{Every frame against every 10th frame.} PA-MPJPE / MPJPE / PVE in mm, on each benchmark's two joint sets (J24: the 24 SMPL joints; J14: the 14 joints of the Human3.6M regressor). $\Delta$PA: PA-MPJPE on every 10th frame minus PA-MPJPE on every frame. Last column: the frames (EMDB-1) or person-frames (3DPW) scored.}
  \label{tab:supp-stride}
\end{table}

\section{Published numbers, under each paper's own protocol}
\label{sec:supp-as-published}
For reference, \cref{tab:supp-as-published} lists the numbers that each paper reports under its own protocol: its own person boxes and joint set, and in some cases its own training data or test-time fitting.
We quote them as printed and never mix them with our rows: they are not comparable with each other, nor with the paper's Tab.~2.
The reproduction gaps in Sec.~4 of the paper compare each method's printed number with our run of its released model, on the joint set of its paper.
The table also covers methods that are not in the paper's Tab.~2, such as video-based methods (WHAM, TRAM, GENMO). SMPLer-X-S32 and Multi-HMR-S get no verdict: no paper reports the variant that is released and timed.

\paragraph{Reproduction gaps.}
On the joint set of each paper, NLF-S, TokenHMR and HMR2.0b reproduce their published numbers within 1\,mm on both EMDB-1 and 3DPW, and CameraHMR on 3DPW.
NLF-L scores 1.9--2.2\,mm better than its paper, and SAM~3D~Body, Fast SAM~3D~Body, HybrIK, MeshPose (3DPW) and CameraHMR on EMDB-1 score 1.3--4.5\,mm worse. MeshPose is compared on 3DPW only, the one benchmark its paper reports, and on its own 14 LSP joints, so its 2.9\,mm gap is approximate.
On RICH, CameraHMR reproduces its published number (0.4\,mm better), SAM~3D~Body scores 2.6\,mm worse and Fast SAM~3D~Body 1.7\,mm better than their papers.
HMR2.0b scores 9.9\,mm worse than the run of its public checkpoint reported by SAM~3D~Body, whose crop is not stated, although it reproduces that same run on 3DPW (0.2\,mm better) and EMDB-1 (0.8\,mm better).

\begin{table}[h]
  \centering
  \footnotesize
  \setlength{\tabcolsep}{4pt}

  \begingroup
  \let\origtabular\tabular
  \renewcommand{\tabular}[1]{\origtabular{@{}>{\raggedright\arraybackslash}p{2.1cm}cc>{\raggedright\arraybackslash}p{4.4cm}>{\raggedright\arraybackslash}p{5.0cm}@{}}}%
  \input{tables/supp_as_published}
  \endgroup
  \caption{\textbf{Published numbers, as printed.} PA-MPJPE / MPJPE in mm, each under its paper's own protocol, with the table it comes from. These rows are not comparable with each other or with the paper's Tab.~2. --: not reported.}
  \label{tab:supp-as-published}
\end{table}

\section{The \method family with detector boxes}
\label{sec:supp-det}
\begin{table}[h]
  \centering
  \resizebox{\linewidth}{!}{\input{tables/supp_family}}
  \caption{\textbf{The \method family, ground-truth boxes against detector boxes.} ``oracle'': ground-truth boxes, the protocol of the paper's Tab.~2; ``matched'': each annotated person takes the box of an off-the-shelf detector (YOLO11n-pose) that overlaps it most (IoU\,$>$\,0.1), or the ground-truth box when none does. \emph{M-C}: the medium architecture trained only on commercially usable images (Harmony4D and COCO). PA-MPJPE and MPJPE in mm; latency in ms (TensorRT fp16 on the RTX 4070, NPU fp16 on the S23).}
  \label{tab:supp-family}
\end{table}

\paragraph{Why detector boxes score better.}
The boxes of the detector are tighter than the ground-truth boxes: the median ratio of their sides is 0.860 on EMDB-1 and 0.852 on 3DPW.
This is why \method-M scores better with them (46.9 against 48.8\,mm PA-MPJPE on EMDB-1, 39.3 against 40.0 on 3DPW, \cref{tab:supp-family}): scaling the ground-truth box by 0.9 before cropping gives 46.6 and 39.2\,mm, within 0.3\,mm of the detector boxes.
The cause is not a difference between box conventions: the box side divided by the extent of the body keypoints is 1.29 on the training labels (a Harmony4D sample), 1.28 on 3DPW and 1.31 on EMDB-1.
It is a property of \method-M, which prefers a tighter framing; the paper therefore treats the crop scale as part of its preprocessing (Sec.~3.3).

\paragraph{The kinematic fit with detector boxes.}
The crop scales of the fit assume the ground-truth box convention, so our pipeline enlarges detector boxes by $1/0.86$ before cropping.
With the matched YOLO11n-pose boxes of \cref{tab:supp-family}, enlarged in this way, the two-crop fit scores 45.1\,mm PA-MPJPE on EMDB-1 and 39.4\,mm on 3DPW, against 44.4 and 39.3\,mm with the ground-truth boxes: the gain holds with detector boxes.
Without the enlargement, it scores 44.5 and 40.1\,mm.
Against \method-M on the same detector boxes (46.9 and 39.3\,mm), the fit gains 1.8\,mm on EMDB-1 and nothing on 3DPW, as in the ablation of the paper (Tab.~4).

\section{Design ablation under the paper's protocol}
\label{sec:supp-design}
Each row of \cref{tab:supp-design} is a training run, scored like every row of the paper's Tab.~2 (ground-truth boxes, calibrated focal, official MHR-to-SMPL conversion, every 10th frame). Errors in mm. The crop-framing runs are complete (600 epochs). The 224\,px runs of the input-size block were stopped at 440--532 epochs, once their validation error had plateaued, so \cref{sec:supp-history} also compares the two input sizes at matched epochs.

\begin{table}[h]
  \centering
  \input{tables/supp_design_ablation}
  \caption{\textbf{Design ablation.} Crop framing: the share of training crops that use the standard crop, the rest being the three other framings (zoomed out, zoomed in, shifted). Input size: an earlier version of the training recipe at four backbone sizes, which is why its B4 288\,px row differs from \method-M. Blue: \method-M.}
  \label{tab:supp-design}
\end{table}

\section{Design results from the training histories}
\label{sec:supp-history}
Two results only exist in the training histories: the 224\,px runs stopped before 600 epochs, and the checkpoints of the linear-head runs were not kept. \Cref{tab:supp-history} gives them on the metric logged during training (3DPW validation, 14 joints through a linear MHR-to-J14 adapter), which is not comparable with the tables scored under the paper's protocol. At matched epochs, 288\,px beats 224\,px at every backbone size, by 1.8--4.0\,mm.

\begin{table}[h]
  \centering
  \resizebox{\linewidth}{!}{\input{tables/supp_history_ablation}}
  \caption{\textbf{Input size at matched epochs and the continuous head.} PA-MPJPE in mm, 3DPW validation. ``288\,px best by then'': the 288\,px run's best score up to the epoch at which its 224\,px twin stopped. The continuous-head runs also enable the head's root-rotation loss and the exact keypoint mapping, which on its own changes the metric by 0.13\,mm.}
  \label{tab:supp-history}
\end{table}

\section{Kinematic fit}
\label{sec:supp-fit}
This section gives the equations behind Sec.~3.3 of the paper.

\paragraph{Unknowns and residuals.}
The fit moves $\mathbf{u}=(\boldsymbol{\theta}_{3:68},\mathbf{t})\in\mathbb{R}^{68}$: the root rotation and the 62 body angles of the rig parameters, and the camera translation.
The bone scales, the identity coefficients $\boldsymbol{\beta}$, the hands and the face keep the network's prediction.
For each of the 30 body keypoints $k$ and each image coordinate $c\in\{x,y\}$, the residual is
\begin{equation}
  r_{k,c}(\mathbf{u}) = \frac{\pi_c\big(\mathbf{J}_k(\mathbf{u})+\mathbf{t}\big) - x_{k,c}}{\sigma_{k,c}}, \qquad
  \pi(\mathbf{X}) = f\,\frac{(X_1, X_2)}{X_3} + (c_x, c_y),
\end{equation}
where $\mathbf{J}_k$ is the keypoint read from the rig's forward kinematics, $x_{k,c}$ the SimCC expectation, and $\sigma_{k,c}$ the standard deviation of the SimCC distribution $\mathbf{P}_{k,c}$, floored at 1\,\% of the crop side.
All 2D quantities are expressed in units of the crop side, centred on the crop.
The fit minimises the loss
\begin{equation}
  \mathcal{L}_{\mathrm{fit}}(\mathbf{u}) = \lVert \mathbf{r}(\mathbf{u}) \rVert^2 + (\mathbf{u}-\mathbf{u}_0)^{\top}\boldsymbol{\Lambda}(\mathbf{u}-\mathbf{u}_0),
\end{equation}
with $\boldsymbol{\Lambda}$ diagonal: 30 for the 65 angles (radians) and 0.01 for the translation (metres).

\paragraph{Exact Jacobian.}
The 65 parameters act only on joint rotations.
If an angle of joint $j$ rotates about the world axis $\mathbf{w}$, its derivative moves every joint $p$ below $j$ by $\mathbf{w}\times(\mathbf{p}-\mathbf{p}_j)$, and the keypoints are a fixed linear map of the joint positions, so
\begin{equation}
  \frac{\partial \mathbf{J}_k}{\partial \alpha} = \sum_{p\,\in\,\mathrm{desc}(j)} A_{k,p}\; \mathbf{w}\times(\mathbf{p}-\mathbf{p}_j),
\end{equation}
where $A$ is the keypoint regressor.
The axes $\mathbf{w}$ follow from the global joint rotations and the rig's Euler order, and the chain rule through $\pi$ gives the $60\times68$ Jacobian $\mathbf{J}_r$ of the residuals.
All of it is batched matrix products over the 127 joints, with no loop over the skeleton.

\paragraph{One Levenberg-Marquardt step.}
From $\mathbf{u}_0$, the network's prediction,
\begin{equation}
  \mathbf{H} = \mathbf{J}_r^{\top}\mathbf{J}_r + \boldsymbol{\Lambda} + \mu\,\mathrm{diag}(\mathbf{J}_r^{\top}\mathbf{J}_r), \qquad
  \mathbf{u}_1 = \mathbf{u}_0 - \mathbf{H}^{-1}\mathbf{J}_r^{\top}\mathbf{r}(\mathbf{u}_0),
\end{equation}
with $\mu=10^{-3}$ (the prior's gradient is zero at $\mathbf{u}_0$).
ONNX has no operator to solve a linear system, so we invert the Jacobi-scaled matrix $\tilde{\mathbf{H}}=\mathbf{D}\mathbf{H}\mathbf{D}$, $\mathbf{D}=\mathrm{diag}(\mathbf{H})^{-1/2}$, by 10 Newton-Schulz iterations,
\begin{equation}
  \mathbf{X}_0 = \mathbf{I}/\lVert\tilde{\mathbf{H}}\rVert_\infty, \qquad \mathbf{X}_{i+1} = \mathbf{X}_i(2\mathbf{I}-\tilde{\mathbf{H}}\mathbf{X}_i),
\end{equation}
and take $\mathbf{H}^{-1}\approx\mathbf{D}\mathbf{X}_{10}\mathbf{D}$.
A step that contains a NaN is dropped, so the output falls back to the network's prediction.
A second step gains about 0.3\,mm on RICH with two crops and costs a second pass through the forward kinematics; we keep one.

\paragraph{fp16.}
The graph runs in fp16 on the NPU, whose largest value is 65,504.
Two choices keep every intermediate value in range.
First, the 2D quantities are in units of the crop side rather than pixels: in pixels, a broad SimCC distribution on a 2,000\,px crop gives $\sigma^2\approx10^5$, and coordinates near 4,000\,px keep only 2\,px of precision.
Second, the residuals and the Jacobian are divided by 16 and the prior by 256, which leaves the minimiser unchanged: with the 1\,\% floor on $\sigma$, the weighted Jacobian reaches about 100 per entry, and $\mathbf{J}_r^{\top}\mathbf{J}_r$ would overflow unscaled.

\paragraph{Crop scale.}
\method-M is more accurate when the person fills more of the crop, probably because a quarter of its training crops are zoomed in, zoomed out or shifted, so that people appear larger in training, on average, than in the standard crop.
For the first crop, we compared box scales of 1, 0.9 and 0.8 on 3DPW validation, which gave 41.6, 40.1 and 41.3\,mm PA-MPJPE, and kept 0.9.
These validation numbers use a faster approximation of our protocol, as does every choice made on validation data.
The second factor, 0.7, was chosen in the same way, on 3DPW validation and 3DPW-train.

\paragraph{Two crops.}
The second crop scales the box by 0.7 instead of 0.9 and goes through the network in the same batch.
Its 2D estimates are converted to the coordinates of the first crop, so that each keypoint has two estimates, $\mathbf{x}^{(1)}$ and $\mathbf{x}^{(2)}$, with spreads $\sigma_{(1)}$ and $\sigma_{(2)}$.
We average them with weights $1/\sigma^2$, so that the more confident estimate counts more,
\begin{equation}
  \mathbf{x} = \frac{\mathbf{x}^{(1)}/\sigma_{(1)}^2+\mathbf{x}^{(2)}/\sigma_{(2)}^2}{1/\sigma_{(1)}^2+1/\sigma_{(2)}^2},
\end{equation}
and a keypoint that falls outside the second crop keeps its first estimate.
The fit starts from the body and the translation predicted on the first crop.

\paragraph{The value of better 2D keypoints.}
The fit can only be as good as its 2D targets.
On two EgoHumans sequences, we fit to simulated targets that keep a fraction $\rho$ of the 2D head's error, $\mathbf{x}^{\star}+\rho(\mathbf{x}-\mathbf{x}^{\star})$ (\cref{tab:supp-oracle}).
Against no fit, the fit gains 0.4 and 0.7\,mm PA-MPJPE with the network's own 2D keypoints ($\rho=1$), but would gain 17.1 and 6.1\,mm with the ground-truth ones ($\rho=0$).

\begin{table}[h]
  \centering
  \begin{tabular}{@{}l ccccc c@{}}
    \toprule
    Share of the 2D head's error kept, $\rho$ & 100\,\% & 75\,\% & 50\,\% & 25\,\% & 0\,\% & no fit \\
    \midrule
    EgoHumans lego 001 & 52.5 & 48.2 & 43.0 & 37.3 & 35.8 & 52.9 \\
    EgoHumans tennis 003 & 33.9 & 32.5 & 30.7 & 28.5 & 28.5 & 34.6 \\
    \bottomrule
  \end{tabular}
  \caption{\textbf{The fit with better 2D keypoints.} PA-MPJPE (mm, 14 joints, a faster stand-in for the protocol of the paper) of \method-M fitted to 2D targets between the ground truth ($\rho=0$) and its own 2D head ($\rho=1$).}
  \label{tab:supp-oracle}
\end{table}

\paragraph{Learned refiners do not transfer.}
Before the fit, we tried three learned ways to correct \method-M after the network, all trained on data outside the test sets.
A small MLP that corrects the pose, trained on 3DPW-train, improves 3DPW test from 40.0 to 38.8\,mm PA-MPJPE, but degrades EMDB-1 from 48.8 to 50.8\,mm and RICH from 41.7 to 43.8\,mm.
A learned 2D refiner, which re-reads each keypoint from higher-resolution backbone features, cuts the held-out 2D error by 29\,\% and improves 3DPW validation from 40.2 to 38.6\,mm once fitted, but barely moves the test sets (3DPW 39.2 to 39.1\,mm, EMDB-1 46.6 to 46.4\,mm, with the 0.9 crop).
A temporal refiner trained on EgoHumans video degrades EMDB-1 to 53.1\,mm and RICH to 53.5\,mm.
Each of them learns corrections specific to its training data.
The fit has no parameter to learn, and it improves EMDB-1 and RICH, on which nothing was tuned.

\section{Temporal metrics}
\label{sec:supp-temporal}
We run each method on every frame of EMDB-1 and 3DPW, as on a live camera, and measure two quantities on the published joint sets, over runs of consecutive frames of one person, with the joints centred on the mid-hip.
The acceleration error is the mean norm of the difference between the second finite differences of the prediction and of the ground truth (mm/frame$^2$).
The jitter is the mean norm of the third finite difference of the prediction, times the frame rate cubed (10\,m/s$^3$); the ground truth's own jitter is 9.5 on EMDB-1 and 18.8 on 3DPW.
All filters are causal: each frame is filtered from the past frames only, so none adds latency.
We compare three of them.
The $\alpha$-$\beta$ filter ($\alpha=0.8$, $\beta=0.005$) was tuned on 56 EgoHumans sequences and 3DPW-train, and confirmed on the fit's outputs.
The tuned 1 Euro filter runs on the fit's pose angles, with a minimum cut-off of 12\,Hz, a speed coefficient of 0.5 and a derivative cut-off of 1\,Hz, chosen on 3DPW validation and the 56 EgoHumans sequences.
Its 12\,Hz minimum cut-off gives a lag of about 13\,ms (0.4 frame).
The untuned 1 Euro filter is the one of our demo application, applied to the mesh around the body (cut-off 1.0\,Hz, speed coefficient 4) and to its position (0.6\,Hz, 2).
The two tuned filters are equally accurate, and the 1 Euro filter is the smoother: its acceleration error is 10\,\% lower and its jitter 12--13\,\% lower.
RICH is left out: it lists only every 5th frame, so a filter tuned at 30 frames per second does not apply.

\begin{table}[h]
  \centering
  \begin{tabular}{@{}l cc cc@{}}
    \toprule
    & \multicolumn{2}{c}{EMDB-1} & \multicolumn{2}{c}{3DPW} \\
    \cmidrule(lr){2-3}\cmidrule(l){4-5}
    Method & Accel. error & Jitter & Accel. error & Jitter \\
    \midrule
    \method-M & 27.1 & 128 & 23.2 & 107 \\
    \method-M + 1 Euro filter, demo setting & 10.0 & 40 & 8.4 & 27 \\
    \method-M + fit (1 crop) & 24.4 & 115 & 23.3 & 107 \\
    \method-M + fit (1 crop) + $\alpha$-$\beta$ filter & 18.3 & 84 & 18.0 & 79 \\
    \method-M + fit (1 crop) + 1 Euro filter, tuned & 16.4 & 74 & 16.2 & 69 \\
    \method-M + fit (2 crops) & 22.5 & 106 & 22.5 & 103 \\
    \method-M + fit (2 crops) + $\alpha$-$\beta$ filter & 17.0 & 78 & 17.4 & 76 \\
    \method-M + fit (2 crops) + 1 Euro filter, tuned & 15.2 & 68 & 15.7 & 67 \\
    NLF-S & 21.2 & 100 & 18.4 & 82 \\
    NLF-S + 1 Euro filter, demo setting & 8.0 & 31 & 7.5 & 21 \\
    \midrule
    Ground truth & -- & 9.5 & -- & 18.8 \\
    \bottomrule
  \end{tabular}
  \caption{\textbf{Temporal metrics on every frame.} Acceleration error in mm/frame$^2$, jitter in 10\,m/s$^3$, lower is better for both. With either tuned filter, the fit's PA-MPJPE improves by 0.1--0.3\,mm. Two crops / one crop, EMDB-1: 44.2 / 44.9\,mm with $\alpha$-$\beta$ and with the tuned 1 Euro filter (44.4 / 45.2 without filter); 3DPW: 39.1 / 39.2\,mm with $\alpha$-$\beta$ and 39.2 / 39.3\,mm with the tuned 1 Euro filter (39.3 / 39.4 without). The demo setting of the 1 Euro filter costs \method-M 0.7\,mm on EMDB-1 and 1.4\,mm on 3DPW.}
  \label{tab:supp-temporal}
\end{table}

\section{Accuracy against latency on a desktop GPU and on 3DPW}
\label{sec:supp-pareto}
\begin{figure}[t]
  \centering
  \includegraphics[width=\textwidth]{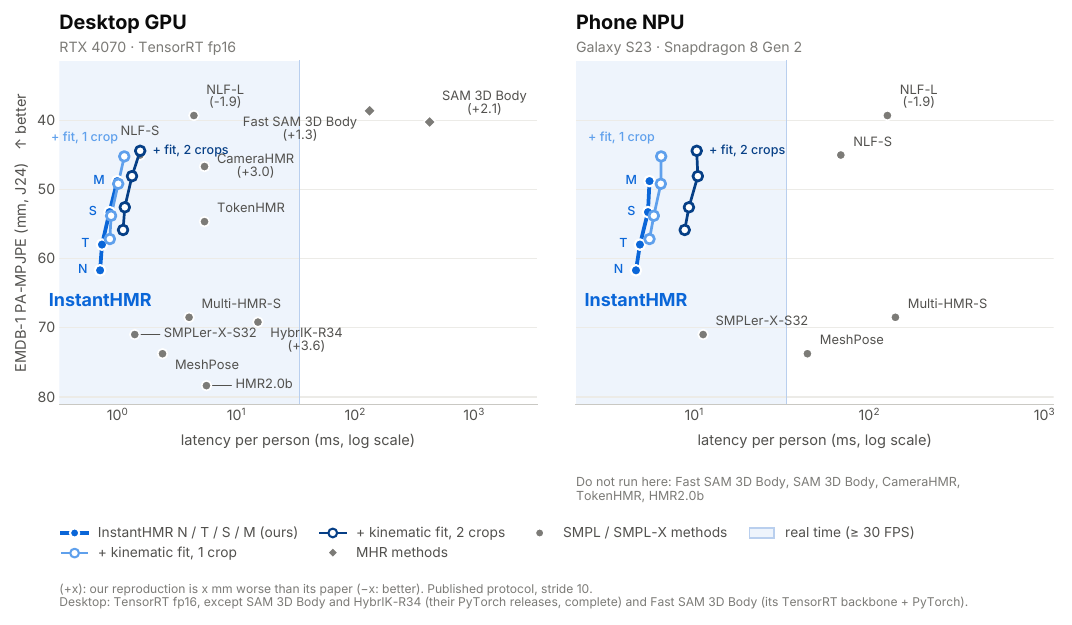}
  \caption{\textbf{Accuracy against latency, on a desktop GPU and on a phone.} EMDB-1 PA-MPJPE (24 joints, lower is better, up) against model latency per person (log scale, faster is left), for every method under the protocol of the paper's Sec.~4. Left: an RTX~4070 with TensorRT fp16 (SAM~3D~Body and HybrIK-R34 in their PyTorch releases, Fast SAM~3D~Body with its TensorRT backbone and the rest in PyTorch). Right: the Samsung Galaxy S23 NPU, as in Fig.~1 of the paper. \method's four sizes form three blue curves: the network alone (filled circles), and with the kinematic fit on one crop (light) and on two crops (dark). The \colorbox{ihmrzone}{shaded zone} is real time ($\geq$\,30\,FPS). (+$x$): our reproduction is $x$\,mm worse than its paper. On the desktop GPU, nearly every method is real time, and the fit costs 0.14--0.16\,ms per size with one crop; on the phone NPU, almost no other method is real time.}
  \label{fig:supp-pareto-emdb}
\end{figure}

\begin{figure}[t]
  \centering
  \includegraphics[width=\textwidth]{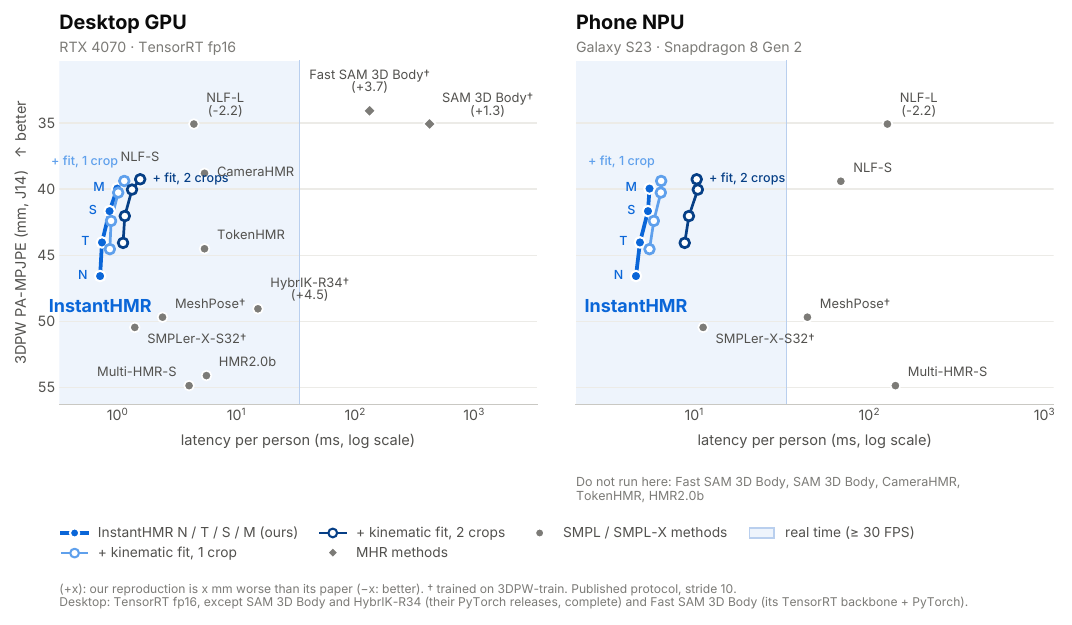}
  \caption{\textbf{Accuracy against latency on 3DPW.} As \cref{fig:supp-pareto-emdb}, with 3DPW PA-MPJPE on 14 joints. The three blue curves are \method's four sizes: the network alone (filled circles), and with the kinematic fit on one crop (light) and on two crops (dark). On 3DPW, the fit gains less than on EMDB-1: 0.6--2.1\,mm with one crop and 0.7--2.5\,mm with two, the smallest sizes gaining most. $^\dagger$: trained on 3DPW-train.}
  \label{fig:supp-pareto-3dpw}
\end{figure}

\Cref{fig:supp-pareto-emdb} adds a desktop-GPU panel to Fig.~1 of the paper, and \cref{fig:supp-pareto-3dpw} draws both panels on 3DPW.

\section{Partial timings}
\label{sec:supp-partial}
The paper's Tab.~1 leaves a cell empty (--) when a method cannot be timed as a whole on that target.
\Cref{tab:supp-partial} gives what we could time there: a part of the method only, so each number is a lower bound ($\geq$) on the method's latency.
HybrIK's network exports, but its analytical inverse kinematics (an SVD and a determinant per joint) and the SMPL forward pass do not; its network alone already takes at least 73.97\,ms on the S23 NPU, 13$\times$ \method-M.
For CameraHMR, TokenHMR and HMR2.0b, only the ViT-H backbone runs on the S23 CPU, where it takes more than 1.8\,s; it does not run on either phone NPU or on the A14 CPU.
SAM~3D~Body and Fast SAM~3D~Body share a DINOv3 ViT-H+ backbone, which takes at least 3.1\,s on the desktop CPU and 14.4\,s on the S23 CPU.
The last two rows break NLF down: its network alone, which predicts 1,024 vertices and 24 joints, takes 8.13\,ms (NLF-S) on the S23 NPU, against 67.97\,ms for the 6,890-vertex graph timed in the paper.
On the NPU, NLF's cost therefore grows with the number of points it predicts, as discussed in the paper's Sec.~5.

\begin{table}[h]
  \centering
  \small
  \resizebox{\linewidth}{!}{%
  \begin{tabular}{@{}l l rrr r rr rr@{}}
    \toprule
    & & \multicolumn{3}{c}{RTX 4070} & CPU & \multicolumn{2}{c}{Phone NPU (fp16)} & \multicolumn{2}{c}{Phone CPU (fp32)} \\
    \cmidrule(lr){3-5}\cmidrule(lr){6-6}\cmidrule(lr){7-8}\cmidrule(l){9-10}
    Method & Timed part & TRT fp16 & ORT fp32 & PyTorch & i5 ORT & S23 & S24 & S23 & A14 \\
    \midrule
    HybrIK-R34 & network (no IK, no SMPL) & $\geq$\,0.69 & $\geq$\,16.48 & & $\geq$\,39.49 & $\geq$\,73.97 & $\geq$\,52.64 & $\geq$\,121 & $\geq$\,385 \\
    CameraHMR & ViT-H backbone & & & & & $\times$ & $\times$ & $\geq$\,1,876 & $\times$ \\
    TokenHMR & ViT-H backbone & & & & & $\times$ & $\times$ & $\geq$\,1,932 & $\times$ \\
    HMR2.0b & ViT-H backbone & & & & & $\times$ & $\times$ & $\geq$\,1,849 & $\times$ \\
    SAM 3D Body, Fast SAM 3D Body & DINOv3 backbone & & & & $\geq$\,3,104 & $\times$ & $\times$ & $\geq$\,14,449 & $\times$ \\
    \midrule
    NLF-S & network (1,024 v + 24 j) & & & 5.10 & & 8.13 & 5.79 & 68.87 & 212.71 \\
    NLF-L & network (1,024 v + 24 j) & & & 10.44 & & 24.83 & 18.74 & 585.32 & 1,821.60 \\
    \bottomrule
  \end{tabular}}
  \caption{\textbf{Partial timings.} Latency per person in ms, on the targets and with the settings of the paper's Tab.~1. Each row times only the part of the method named in the second column. Empty: the whole method is timed in the paper's Tab.~1, or the part was not timed there. $\times$: does not run on the device. The NLF rows leave out the SMPL fit and the SMPL forward pass of NLF's own recipe; they are a breakdown, not the rows of the paper.}
  \label{tab:supp-partial}
\end{table}

\section{On-device results}
\label{sec:supp-device}
This section details the Android app of the paper's Sec.~5.

\paragraph{Setup.}
The app runs \method-M with the one-crop fit on a Samsung Galaxy S23 (Snapdragon 8 Gen 2, Hexagon NPU v73, Android 16).
It uses ONNX Runtime 1.29 with its QNN execution provider (QNN 2.42), in fp16, with the NPU in burst mode and the session at high context priority.
At normal priority, about 40\,\% of the model runs took about 15\,ms instead of 7, most likely because another client of the NPU preempted them; at high priority, 1--2\,\% do.
CPU fallback is disabled, so a model loads only if every one of its layers runs on the NPU.
The model is the graph timed on AI Hub in the paper's Tab.~1 (6.40\,ms), with the outputs the app needs: the 2D keypoints and their confidence, the 3D keypoints, and MHR's 595-vertex mesh (LOD~6) instead of the 6,890-vertex mesh.
The person detector is NanoDet-Plus-m (416$\times$416 input), also entirely on the NPU.
The app tracks one person. The detector finds the person when tracking starts, and is re-run once per second.
On the other frames, the crop comes from the previous frame's body keypoints, enlarged by the margins of the training boxes.
The 288\,px crop is cut on the CPU with four threads.

\paragraph{Timings.}
For a repeatable input, we replay an EMDB video (P7/60\_outdoor\_workout, 720$\times$960, decoded in hardware) in a loop at 30\,FPS for five minutes, after a 10\,s warm-up, with the phone unlocked and on battery.
\Cref{tab:supp-device-latency} gives the time of each stage.
The model takes 1.0\,ms more in the app than on AI Hub, although its mesh is smaller, mostly because the app's timing includes copying the input and output tensors.
A frame without a detection takes 12.73\,ms; a frame with one takes 31.4\,ms, close to the 33.3\,ms that a frame lasts at 30\,FPS.

\begin{table}[h]
  \centering
  \small
  \begin{tabular}{@{}l rr@{}}
    \toprule
    Stage (Galaxy S23, replay, on battery) & Median & 95th pct. \\
    \midrule
    Crop and normalisation, 288\,px (CPU, 4 threads) & 5.21 & 5.62 \\
    Model: network, fit, 3D keypoints, 595-vertex mesh (NPU, with tensor copies) & 7.41 & 7.77 \\
    Whole frame, tracked (8,696 frames) & 12.73 & 13.46 \\
    \midrule
    Detector run: resize, NanoDet on the NPU, NMS (once per second) & 19.6 & 20.8 \\
    Whole frame with a detection (300 frames) & 31.4 & 33.8 \\
    \midrule
    For comparison: the model alone on AI Hub, 6,890-vertex mesh (paper's Tab.~1) & 6.40 & -- \\
    \bottomrule
  \end{tabular}
  \caption{\textbf{Per-stage latency in the Android app}, in ms per frame, over the five-minute replay (8,996 frames). The stages run one after the other; the whole frame also includes the app's bookkeeping.}
  \label{tab:supp-device-latency}
\end{table}

\paragraph{Sustained frame rate.}
\Cref{tab:supp-device-runs} summarises two long runs: the replay above, and a live run with the camera, in which a person 2--3\,m from the phone did squats, arm raises and turns for 342\,s (after a 10\,s warm-up), with the mesh drawn over the image.
In both, the app kept 30.0\,FPS, dropped 1 or 2 frames, tracked the person on every frame, and the phone stayed in its normal thermal state (Android thermal status 0).
In the live run, the median timings per second were close to the replay's: 12.17\,ms per frame, 7.44\,ms for the model and 4.41\,ms for the crop.
There, the 3D keypoints projected into the image lie 2.2\,px from the 2D keypoints (median; 6.3\,px at the 90th percentile), against 15.0\,px for an earlier \method model without the fit: the fit makes the 3D body agree with the 2D keypoints on the phone too.
From the camera sensor to the drawn overlay, a frame takes 124\,ms (median); 94\,ms of it pass inside the camera stack before the app receives the frame, so the app's own processing and drawing take the other 30\,ms.

\begin{table}[h]
  \centering
  \small
  \begin{tabular}{@{}l cc@{}}
    \toprule
    & Replay (EMDB video) & Live camera \\
    \midrule
    Duration & 5\,min & 342\,s \\
    Frame rate & 30.0\,FPS & 30.0\,FPS (10th pct. 29.6, min.\ 29.2) \\
    Frames dropped & 2 of 8,996 & 1 of 10,409 \\
    Person tracked & every frame & all 342\,s \\
    Android thermal status & 0 throughout & 0 throughout \\
    \midrule
    Whole-phone power, mean (idle: 0.79\,W) & 2.46\,W & 4.47\,W \\
    Above idle & +1.66\,W & +3.68\,W \\
    Battery level & 25 $\to$ 23\,\% in 289\,s & 23 $\to$ 20\,\% in 333\,s \\
    \bottomrule
  \end{tabular}
  \caption{\textbf{Sustained runs on the Galaxy S23, on battery.} Power: the whole phone, from the battery's current and voltage sampled once a second; means over the time the phone ran on battery. Idle: screen on, app loaded, nothing running.}
  \label{tab:supp-device-runs}
\end{table}

\paragraph{fp16 on the NPU against fp32.}
On 23 real frames saved by the app, we compare the app's fp16 outputs on the NPU with the fp32 graph run on a desktop CPU, on the same input tensors.
The 3D keypoints differ by 2.93\,mm on average (4.93\,mm at the 95th percentile, 8.14\,mm at most), the mesh vertices by 2.81\,mm (5.06 and 8.25\,mm), and the 2D keypoints by 0.13\,px in the 288\,px crop.
These differences are small against the errors in the paper's Tab.~2 (45.2\,mm PA-MPJPE on EMDB-1 for this model), so we take its accuracy to hold on the phone; we did not score the benchmarks on the phone itself.

\paragraph{Power.}
We read the whole phone's battery current and voltage once a second through Android's battery manager, with the screen brightness fixed.
This includes the screen, the camera, the video decoder and everything else that runs on the phone, so it is an upper bound on what the model draws.
The gauge is noisy from one second to the next, so we quote means (\cref{tab:supp-device-runs}).
During the replay, the phone draws 1.66\,W above idle, about 55\,mJ per frame for the whole pipeline: video decoding, detector, crop, model and drawing.
During the live run, it draws 3.68\,W above idle. Most of the difference with the replay comes from the camera, but we did not measure the camera alone.
These figures have clear limits: one phone, one run of each kind, a gauge sampled once a second, and no camera-only baseline.

%% file: tables/tab_protocol_ablation.tex
\begin{tabular}{@{}lll ccc ccc@{}}
\toprule
 & & & \multicolumn{3}{c}{EMDB-1 (J24, mm)} & \multicolumn{3}{c}{3DPW (J14, mm)} \\
\cmidrule(lr){4-6}\cmidrule(l){7-9}
Boxes & Focal & Conversion & PA & MPJPE & PVE & PA & MPJPE & PVE \\
\midrule
\textcolor{ihmrblue}{oracle} & \textcolor{ihmrblue}{calibrated} & \textcolor{ihmrblue}{Official (Meta)} & \textcolor{ihmrblue}{48.8} & \textcolor{ihmrblue}{74.9} & \textcolor{ihmrblue}{89.1} & \textcolor{ihmrblue}{40.0} & \textcolor{ihmrblue}{64.0} & \textcolor{ihmrblue}{76.8} \\
\midrule
oracle & 55$^\circ$ FOV & Fast's MLP & 65.9 & 161.8 & 189.8 & 43.9 & 81.8 & 95.9 \\
oracle & 55$^\circ$ FOV & J14 adapter & -- & -- & -- & 40.4 & 65.4 & -- \\
oracle & 55$^\circ$ FOV & Official (Meta) & 48.9 & 73.9 & 87.9 & 40.0 & 64.2 & 77.0 \\
oracle & 55$^\circ$ FOV & our SMPL fit & 49.2 & 74.2 & 88.8 & 40.7 & 64.8 & 77.6 \\
oracle & 55$^\circ$ FOV & surface map & 50.1 & 75.0 & 90.0 & 41.4 & 64.9 & 78.1 \\
oracle & calibrated & Fast's MLP & 65.8 & 160.6 & 188.0 & 43.8 & 81.8 & 95.9 \\
oracle & calibrated & J14 adapter & -- & -- & -- & 40.4 & 65.2 & -- \\
oracle & calibrated & our SMPL fit & 49.1 & 75.1 & 90.0 & 40.7 & 64.6 & 77.4 \\
oracle & calibrated & surface map & 49.9 & 75.7 & 91.0 & 41.4 & 64.8 & 77.9 \\
matched & 55$^\circ$ FOV & Fast's MLP & 64.4 & 162.3 & 190.3 & 43.1 & 81.4 & 95.0 \\
matched & 55$^\circ$ FOV & J14 adapter & -- & -- & -- & 39.8 & 64.5 & -- \\
matched & 55$^\circ$ FOV & Official (Meta) & 47.0 & 72.5 & 85.8 & 39.3 & 63.3 & 75.4 \\
matched & 55$^\circ$ FOV & our SMPL fit & 47.3 & 72.8 & 86.7 & 40.0 & 63.9 & 75.9 \\
matched & 55$^\circ$ FOV & surface map & 48.3 & 73.6 & 87.9 & 40.8 & 64.0 & 76.4 \\
matched & calibrated & Fast's MLP & 64.4 & 161.5 & 189.0 & 43.1 & 81.4 & 95.0 \\
matched & calibrated & J14 adapter & -- & -- & -- & 39.8 & 64.4 & -- \\
matched & calibrated & Official (Meta) & 46.9 & 73.5 & 87.1 & 39.3 & 63.1 & 75.2 \\
matched & calibrated & our SMPL fit & 47.3 & 73.8 & 88.0 & 40.0 & 63.8 & 75.7 \\
matched & calibrated & surface map & 48.1 & 74.3 & 88.9 & 40.8 & 63.9 & 76.2 \\
auto & 55$^\circ$ FOV & Fast's MLP & 64.4 & 162.2 & 190.2 & 43.0 & 80.8 & 94.4 \\
auto & 55$^\circ$ FOV & J14 adapter & -- & -- & -- & 39.6 & 64.1 & -- \\
auto & 55$^\circ$ FOV & Official (Meta) & 47.0 & 72.5 & 85.8 & 39.1 & 62.9 & 74.9 \\
auto & 55$^\circ$ FOV & our SMPL fit & 47.3 & 72.8 & 86.7 & 39.8 & 63.5 & 75.4 \\
auto & 55$^\circ$ FOV & surface map & 48.3 & 73.6 & 87.9 & 40.5 & 63.6 & 75.9 \\
auto & calibrated & Fast's MLP & 64.4 & 161.4 & 188.8 & 42.9 & 80.8 & 94.4 \\
auto & calibrated & J14 adapter & -- & -- & -- & 39.6 & 63.9 & -- \\
auto & calibrated & Official (Meta) & 46.9 & 73.5 & 87.0 & 39.1 & 62.7 & 74.7 \\
auto & calibrated & our SMPL fit & 47.2 & 73.8 & 88.0 & 39.8 & 63.3 & 75.2 \\
auto & calibrated & surface map & 48.0 & 74.3 & 88.9 & 40.6 & 63.4 & 75.7 \\
\bottomrule
\end{tabular}

%% file: tables/supp_stride_check.tex
\begin{tabular}{lcccccc}
\toprule
method & benchmark & joint set & every frame & stride 10 & $\Delta$ PA & frames (every / stride 10) \\
\midrule
InstantHMR-M & EMDB-1 & J24 & 48.74 / 74.85 / 89.07 & 48.80 / 74.86 / 89.13 & +0.06 & 24117 / 2419 \\
InstantHMR-M & EMDB-1 & J14 & 44.81 / 73.73 / 86.04 & 44.87 / 73.68 / 86.04 & +0.06 & 24117 / 2419 \\
NLF-S & EMDB-1 & J24 & 45.01 / 74.55 / 87.98 & 45.03 / 74.56 / 87.94 & +0.02 & 24117 / 2419 \\
NLF-S & EMDB-1 & J14 & 41.78 / 71.36 / 83.78 & 41.84 / 71.35 / 83.69 & +0.06 & 24117 / 2419 \\
InstantHMR-M & 3DPW & J14 & 40.23 / 64.36 / 77.21 & 39.96 / 64.01 / 76.82 & -0.27 & 35515 / 3562 \\
InstantHMR-M & 3DPW & J24 & 45.64 / 64.51 / 77.37 & 45.34 / 64.16 / 76.94 & -0.30 & 35515 / 3562 \\
NLF-S & 3DPW & J14 & 39.43 / 62.37 / 74.83 & 39.41 / 62.21 / 74.59 & -0.02 & 35515 / 3562 \\
NLF-S & 3DPW & J24 & 43.42 / 63.73 / 75.56 & 43.38 / 63.56 / 75.31 & -0.04 & 35515 / 3562 \\
\bottomrule
\end{tabular}

%% file: tables/supp_as_published.tex
\begin{tabular}{lcccc}
\toprule
method & 3DPW PA / MPJPE & EMDB-1 PA / MPJPE & source & note \\
\midrule
NLF-S & 38.5 / 60.9 & 44.9 / 74.5 & NLF (NeurIPS'24) Table 2 (trained without 3DPW-train) and Table 3 & Without +fit. With 3DPW-train: 3DPW PA 35.9 / MPJPE 55.6. Which training set the released weights used is not stated. \\
NLF-L & 37.3 / 60.3 & 41.2 / 69.6 & NLF Table 2 (without 3DPW-train) and Table 3 & Without +fit. +fit: EMDB PA 40.9 / MPJPE 68.4. \\
BEDLAM-CLIFF & 46.6 / 72.0 & 60.6 / 98.0 & NLF Table 2 and Table 3 & CameraHMR Table 4 gives EMDB PA 61.3 / MPJPE 97.1. \\
HybrIK-R34 & 44.5 / 72.4 & 65.6 / 103.0 & 3DPW: HybrIK README model zoo (ResNet-34, trained with 3DPW). EMDB-1: NLF Table 3 & The EMDB-1 row's backbone is not stated in NLF and may not be the R34 variant that is timed. \\
PyMAF-R50 & 47.1 / 78.0 & 82.9 / 131.1 & NLF Table 2 and Table 3 &  \\
SMPLer-X-S32 & -- / -- & -- / -- & -- & Only SMPLer-X-H is reported in these papers (SAM 3D Body Table 2: 3DPW PA 46.6, EMDB PA 64.5, public checkpoint). \\
MeshPose & 46.7 / 76.1 & -- / -- & MeshPose (CVPR'24) Table 3 (HRNet-32) & Non-parametric mesh; 14 LSP joints from its own landmarker; trained with 3DPW-train. MeshPose-S / -XS: 3DPW PA 49.0 / 49.7. \\
CameraHMR & 38.5 / 62.1 & 43.7 / 73.0 & CameraHMR (3DV'25) Table 4, default model & Fine-tuned on 3DPW (CameraHMR Table 4, *; also SAM 3D Body Table 2): 3DPW 35.1 / 56.0, EMDB 43.3 / 70.3. RICH PA 34.0 (SAM 3D Body Table 2). \\
HMR2.0b & 54.3 / 81.3 & 79.2 / 118.5 & SAM 3D Body Table 2 and CameraHMR Table 4 & Public checkpoint. RICH PA 48.1 (SAM 3D Body Table 2). \\
PromptHMR & 36.1 / 58.7 & 41.0 / 71.7 & SAM 3D Body Table 2 &  \\
Multi-HMR & 46.9 / 69.5 & -- / -- & NLF Table 2 (without 3DPW-train) & Variant of the cited row not stated; the timed backbone is ViT-S at 672 px. \\
SAM 3D Body & 33.8 / 54.8 & 38.2 / 61.7 & SAM 3D Body Table 2 (3DB-DINOv3) &  \\
Fast SAM 3D Body & 30.4 / 59.5 & 37.3 / 64.3 & Fast SAM 3D Body Table 1 (oracle boxes) & Automatic boxes: 3DPW PA 29.7, EMDB PA 36.2. \\
WHAM & 35.9 / 57.8 & 50.4 / 79.7 & SAM 3D Body Table 2 & Video-based. \\
TRAM & 35.6 / 59.3 & 45.7 / 74.4 & SAM 3D Body Table 2 & Video-based. \\
GENMO & 34.6 / 53.9 & 42.5 / 73.0 & SAM 3D Body Table 2 & Video-based. \\
TokenHMR & 44.8 / 70.9 & 55.6 / 89.5 & TokenHMR README model zoo, TokenHMR-ITW (paper version) & PVE 84.3 / 108.5 \\
\bottomrule
\end{tabular}

%% file: tables/supp_family.tex
\begin{tabular}{@{}l cc cc cc cc cc@{}}
\toprule
 & \multicolumn{4}{c}{EMDB-1 (J24, mm)} & \multicolumn{4}{c}{3DPW (J14, mm)} & \multicolumn{2}{c}{Latency (ms)} \\
\cmidrule(lr){2-5}\cmidrule(lr){6-9}\cmidrule(l){10-11}
 & \multicolumn{2}{c}{oracle} & \multicolumn{2}{c}{matched} & \multicolumn{2}{c}{oracle} & \multicolumn{2}{c}{matched} & & \\
Model & PA & MPJPE & PA & MPJPE & PA & MPJPE & PA & MPJPE & RTX 4070 & S23 NPU \\
\midrule
\method-N (nano) & 61.7 & 90.8 & 59.4 & 89.0 & 46.6 & 73.4 & 45.2 & 71.8 & 0.70 & 4.60 \\
\method-T (tiny) & 58.0 & 85.9 & 56.3 & 84.5 & 44.0 & 68.7 & 43.0 & 67.5 & 0.73 & 4.84 \\
\method-S (small) & 53.3 & 79.9 & 51.4 & 78.4 & 41.7 & 65.9 & 40.3 & 64.3 & 0.84 & 5.38 \\
\textcolor{ihmrblue}{\textbf{\method-M (medium)}} & \textcolor{ihmrblue}{48.8} & \textcolor{ihmrblue}{74.9} & \textcolor{ihmrblue}{46.9} & \textcolor{ihmrblue}{73.5} & \textcolor{ihmrblue}{40.0} & \textcolor{ihmrblue}{64.0} & \textcolor{ihmrblue}{39.3} & \textcolor{ihmrblue}{63.1} & \textcolor{ihmrblue}{0.98} & \textcolor{ihmrblue}{5.49} \\
\midrule
\method-M-C (medium-commercial) & 72.2 & 114.6 & 70.3 & 110.8 & 50.7 & 80.6 & 47.8 & 74.9 & 0.98 & 5.49 \\
\bottomrule
\end{tabular}

%% file: tables/supp_design_ablation.tex
\begin{tabular}{@{}l cc cc@{}}
\toprule
 & \multicolumn{2}{c}{EMDB-1 (J24, mm)} & \multicolumn{2}{c}{3DPW (J14, mm)} \\
\cmidrule(lr){2-3}\cmidrule(l){4-5}
Training run & PA & MPJPE & PA & MPJPE \\
\midrule
\multicolumn{5}{@{}l}{\textit{Crop framing (HGNetV2-B4, 288 px)}} \\
standard crop only & 58.8 & 82.1 & 42.7 & 64.4 \\
standard crop 50 \%, other framings 50 \% & 50.4 & 77.3 & 40.8 & 64.5 \\
\textcolor{ihmrblue}{\textbf{standard crop 75 \% (\method-M)}} & \textcolor{ihmrblue}{48.8} & \textcolor{ihmrblue}{74.9} & \textcolor{ihmrblue}{40.0} & \textcolor{ihmrblue}{64.0} \\
four framings, uniform & 51.2 & 78.4 & 41.1 & 65.8 \\
\midrule
\multicolumn{5}{@{}l}{\textit{Input size (earlier training recipe)}} \\
HGNetV2-B0, 224 px & 67.0 & 96.6 & 50.1 & 78.1 \\
HGNetV2-B0, 288 px & 61.5 & 89.3 & 47.2 & 73.7 \\
HGNetV2-B1, 224 px & 64.0 & 92.5 & 48.4 & 75.1 \\
HGNetV2-B1, 288 px & 59.5 & 88.2 & 46.0 & 71.7 \\
HGNetV2-B2, 224 px & 59.4 & 86.9 & 45.9 & 71.6 \\
HGNetV2-B2, 288 px & 55.5 & 82.4 & 43.2 & 68.0 \\
HGNetV2-B4, 224 px & 54.2 & 81.1 & 43.9 & 70.0 \\
HGNetV2-B4, 288 px & 49.8 & 76.0 & 42.2 & 67.0 \\
\bottomrule
\end{tabular}

%% file: tables/supp_history_ablation.tex
\begin{tabular}{@{}l cccc@{}}
\toprule
\multicolumn{5}{@{}l}{\textit{Input size at matched epochs (earlier training recipe; 224\,px runs stopped early)}} \\
Backbone & 224\,px stopped at & 224\,px best & 288\,px best by then & 288\,px best (600) \\
\midrule
HGNetV2-B0 & 532 & 52.5 & 48.5 & 48.5 \\
HGNetV2-B1 & 500 & 49.3 & 47.0 & 47.0 \\
HGNetV2-B2 & 440 & 46.7 & 44.9 & 44.9 \\
HGNetV2-B4 & 449 & 43.4 & 41.2 & 40.8 \\
\midrule
\multicolumn{5}{@{}l}{\textit{Continuous MHR head (HGNetV2-B4, 300 epochs, two seeds)}} \\
Run & \multicolumn{2}{c}{best (seed 0 / 1)} & \multicolumn{2}{c}{mean of last 20 epochs (seed 0 / 1)} \\
\midrule
linear head & \multicolumn{2}{c}{41.92 / 41.75} & \multicolumn{2}{c}{42.07 / 41.81} \\
continuous head + root-rotation loss + exact keypoint mapping & \multicolumn{2}{c}{40.64 / 40.57} & \multicolumn{2}{c}{40.66 / 40.75} \\
\bottomrule
\end{tabular}

%% file: main.bbl
\begin{thebibliography}{54}
\providecommand{\natexlab}[1]{#1}
\providecommand{\url}[1]{\texttt{#1}}
\expandafter\ifx\csname urlstyle\endcsname\relax
  \providecommand{\doi}[1]{doi: #1}\else
  \providecommand{\doi}{doi: \begingroup \urlstyle{rm}\Url}\fi

\bibitem[Andriluka et~al.(2014)Andriluka, Pishchulin, Gehler, and
  Schiele]{andriluka2014mpii}
Mykhaylo Andriluka, Leonid Pishchulin, Peter Gehler, and Bernt Schiele.
\newblock {2D} human pose estimation: New benchmark and state of the art
  analysis.
\newblock In \emph{CVPR}, 2014.

\bibitem[Baradel et~al.(2024)Baradel, Armando, Galaaoui, Br{\'e}gier,
  Weinzaepfel, Rogez, and Lucas]{baradel2024multihmr}
Fabien Baradel, Matthieu Armando, Salma Galaaoui, Romain Br{\'e}gier, Philippe
  Weinzaepfel, Gr{\'e}gory Rogez, and Thomas Lucas.
\newblock {Multi-HMR}: Multi-person whole-body human mesh recovery in a single
  shot.
\newblock In \emph{ECCV}, 2024.

\bibitem[Bogo et~al.(2016)Bogo, Kanazawa, Lassner, Gehler, Romero, and
  Black]{bogo2016smplify}
Federica Bogo, Angjoo Kanazawa, Christoph Lassner, Peter Gehler, Javier Romero,
  and Michael~J. Black.
\newblock Keep it {SMPL}: Automatic estimation of {3D} human pose and shape
  from a single image.
\newblock In \emph{ECCV}, 2016.

\bibitem[Br{\'e}gier et~al.(2025)Br{\'e}gier, Fiche, Bravo-S{\'a}nchez, Lucas,
  Armando, Weinzaepfel, Rogez, and Baradel]{bregier2025anny}
Romain Br{\'e}gier, Gu{\'e}nol{\'e} Fiche, Laura Bravo-S{\'a}nchez, Thomas
  Lucas, Matthieu Armando, Philippe Weinzaepfel, Gr{\'e}gory Rogez, and Fabien
  Baradel.
\newblock Human mesh modeling for {Anny} body.
\newblock \emph{arXiv preprint arXiv:2511.03589}, 2025.

\bibitem[Cai et~al.(2023)Cai, Yin, Zeng, Wei, Sun, Wang, Pang, Mei, Zhang,
  Zhang, Loy, Yang, and Liu]{cai2023smplerx}
Zhongang Cai, Wanqi Yin, Ailing Zeng, Chen Wei, Qingping Sun, Yanjun Wang,
  Hui~En Pang, Haiyi Mei, Mingyuan Zhang, Lei Zhang, Chen~Change Loy, Lei Yang,
  and Ziwei Liu.
\newblock {SMPLer-X}: Scaling up expressive human pose and shape estimation.
\newblock In \emph{NeurIPS Datasets and Benchmarks}, 2023.

\bibitem[Carion et~al.(2020)Carion, Massa, Synnaeve, Usunier, Kirillov, and
  Zagoruyko]{carion2020detr}
Nicolas Carion, Francisco Massa, Gabriel Synnaeve, Nicolas Usunier, Alexander
  Kirillov, and Sergey Zagoruyko.
\newblock End-to-end object detection with transformers.
\newblock In \emph{ECCV}, 2020.

\bibitem[Casiez et~al.(2012)Casiez, Roussel, and Vogel]{casiez2012oneeuro}
G{\'e}ry Casiez, Nicolas Roussel, and Daniel Vogel.
\newblock {1 Euro} filter: A simple speed-based low-pass filter for noisy input
  in interactive systems.
\newblock In \emph{CHI}, 2012.

\bibitem[Dwivedi et~al.(2024)Dwivedi, Sun, Patel, Feng, and
  Black]{dwivedi2024tokenhmr}
Sai~Kumar Dwivedi, Yu Sun, Priyanka Patel, Yao Feng, and Michael~J. Black.
\newblock {TokenHMR}: Advancing human mesh recovery with a tokenized pose
  representation.
\newblock In \emph{CVPR}, 2024.

\bibitem[Ferguson et~al.(2025)Ferguson, Osman, Bescos, Stoll, Twigg,
  et~al.]{ferguson2025mhr}
Aaron Ferguson, Ahmed A.~A. Osman, Berta Bescos, Carsten Stoll, Chris Twigg,
  et~al.
\newblock {MHR}: Momentum human rig.
\newblock \emph{arXiv preprint arXiv:2511.15586}, 2025.

\bibitem[Goel et~al.(2023)Goel, Pavlakos, Rajasegaran, Kanazawa, and
  Malik]{goel2023hmr2}
Shubham Goel, Georgios Pavlakos, Jathushan Rajasegaran, Angjoo Kanazawa, and
  Jitendra Malik.
\newblock Humans in {4D}: Reconstructing and tracking humans with transformers.
\newblock In \emph{ICCV}, 2023.

\bibitem[Grishchenko et~al.(2022)Grishchenko, Bazarevsky, Zanfir, Bazavan,
  Zanfir, Yee, Raveendran, Zhdanovich, Grundmann, and
  Sminchisescu]{grishchenko2022blazeposeghum}
Ivan Grishchenko, Valentin Bazarevsky, Andrei Zanfir, Eduard~Gabriel Bazavan,
  Mihai Zanfir, Richard Yee, Karthik Raveendran, Matsvei Zhdanovich, Matthias
  Grundmann, and Cristian Sminchisescu.
\newblock {BlazePose GHUM Holistic}: Real-time {3D} human landmarks and pose
  estimation.
\newblock \emph{arXiv preprint arXiv:2206.11678}, 2022.

\bibitem[He et~al.(2024)He, Luo, Xiao, Zhang, Kitani, Liu, and Shi]{he2024h2o}
Tairan He, Zhengyi Luo, Wenli Xiao, Chong Zhang, Kris Kitani, Changliu Liu, and
  Guanya Shi.
\newblock Learning human-to-humanoid real-time whole-body teleoperation.
\newblock In \emph{IROS}, 2024.

\bibitem[Huang et~al.(2022)Huang, Yi, H{\"o}schle, Safroshkin, Alexiadis,
  Polikovsky, Scharstein, and Black]{huang2022rich}
Chun-Hao~P. Huang, Hongwei Yi, Markus H{\"o}schle, Matvey Safroshkin,
  Tsvetelina Alexiadis, Senya Polikovsky, Daniel Scharstein, and Michael~J.
  Black.
\newblock Capturing and inferring dense full-body human-scene contact.
\newblock In \emph{Proceedings of the IEEE/CVF Conference on Computer Vision
  and Pattern Recognition (CVPR)}, pages 13274--13285, 2022.

\bibitem[Jiang et~al.(2022)Jiang, Streli, Qiu, Fender, Laich, Snape, and
  Holz]{jiang2022avatarposer}
Jiaxi Jiang, Paul Streli, Huajian Qiu, Andreas Fender, Larissa Laich, Patrick
  Snape, and Christian Holz.
\newblock {AvatarPoser}: Articulated full-body pose tracking from sparse motion
  sensing.
\newblock In \emph{European Conference on Computer Vision (ECCV)}, pages
  443--460, 2022.

\bibitem[Jiao et~al.(2024)Jiao, Huang, Wang, Jia, Zhong, and Cai]{jiao2024reba}
Zeyu Jiao, Kai Huang, Qun Wang, Guozhu Jia, Zhenyu Zhong, and Yingjie Cai.
\newblock Improved {REBA}: deep learning based rapid entire body risk
  assessment for prevention of musculoskeletal disorders.
\newblock \emph{Ergonomics}, 2024.

\bibitem[Joo et~al.(2021)Joo, Neverova, and Vedaldi]{joo2021eft}
Hanbyul Joo, Natalia Neverova, and Andrea Vedaldi.
\newblock Exemplar fine-tuning for {3D} human model fitting towards in-the-wild
  {3D} human pose estimation.
\newblock In \emph{3DV}, 2021.

\bibitem[Kanazawa et~al.(2018)Kanazawa, Black, Jacobs, and
  Malik]{kanazawa2018hmr}
Angjoo Kanazawa, Michael~J. Black, David~W. Jacobs, and Jitendra Malik.
\newblock End-to-end recovery of human shape and pose.
\newblock In \emph{CVPR}, 2018.

\bibitem[Kaufmann et~al.(2023)Kaufmann, Song, Guo, Shen, Jiang, Tang,
  Z{\'a}rate, and Hilliges]{kaufmann2023emdb}
Manuel Kaufmann, Jie Song, Chen Guo, Kaiyue Shen, Tianjian Jiang, Chengcheng
  Tang, Juan~Jos{\'e} Z{\'a}rate, and Otmar Hilliges.
\newblock {EMDB}: The electromagnetic database of global {3D} human pose and
  shape in the wild.
\newblock In \emph{ICCV}, 2023.

\bibitem[Khirodkar et~al.(2024)Khirodkar, Song, Cao, Luo, and
  Kitani]{khirodkar2024harmony4d}
Rawal Khirodkar, Jyun-Ting Song, Jinkun Cao, Zhengyi Luo, and Kris Kitani.
\newblock {Harmony4D}: A video dataset for in-the-wild close human
  interactions.
\newblock In \emph{NeurIPS Datasets and Benchmarks}, 2024.

\bibitem[Kirillov et~al.(2023)Kirillov, Mintun, Ravi, Mao, Rolland, Gustafson,
  Xiao, Whitehead, Berg, Lo, Doll{\'a}r, and Girshick]{kirillov2023sam}
Alexander Kirillov, Eric Mintun, Nikhila Ravi, Hanzi Mao, Chloe Rolland, Laura
  Gustafson, Tete Xiao, Spencer Whitehead, Alexander~C. Berg, Wan-Yen Lo, Piotr
  Doll{\'a}r, and Ross Girshick.
\newblock Segment anything.
\newblock In \emph{ICCV}, 2023.

\bibitem[Kolotouros et~al.(2019)Kolotouros, Pavlakos, Black, and
  Daniilidis]{kolotouros2019spin}
Nikos Kolotouros, Georgios Pavlakos, Michael~J. Black, and Kostas Daniilidis.
\newblock Learning to reconstruct {3D} human pose and shape via model-fitting
  in the loop.
\newblock In \emph{ICCV}, 2019.

\bibitem[L{\^e} et~al.(2024)L{\^e}, Kakolyris, Koutras, Tam, Skordos,
  Papandreou, G{\"u}ler, and Kokkinos]{le2024meshpose}
Eric-Tuan L{\^e}, Antonis Kakolyris, Petros Koutras, Himmy Tam, Efstratios
  Skordos, George Papandreou, R{\i}za~Alp G{\"u}ler, and Iasonas Kokkinos.
\newblock {MeshPose}: Unifying {DensePose} and {3D} body mesh reconstruction.
\newblock In \emph{CVPR}, 2024.

\bibitem[Li et~al.(2021)Li, Xu, Chen, Bian, Yang, and Lu]{li2021hybrik}
Jiefeng Li, Chao Xu, Zhicun Chen, Siyuan Bian, Lixin Yang, and Cewu Lu.
\newblock {HybrIK}: A hybrid analytical-neural inverse kinematics solution for
  {3D} human pose and shape estimation.
\newblock In \emph{CVPR}, 2021.

\bibitem[Li et~al.(2025)Li, Cao, Zhang, Rempe, Kautz, Iqbal, and
  Yuan]{li2025gem}
Jiefeng Li, Jinkun Cao, Haotian Zhang, Davis Rempe, Jan Kautz, Umar Iqbal, and
  Ye Yuan.
\newblock {GEM}: A generalist model for human motion.
\newblock In \emph{ICCV}, 2025.

\bibitem[Li et~al.(2020)Li, Martin, and Xu]{li2020rula}
Li Li, Tara Martin, and Xu Xu.
\newblock A novel vision-based real-time method for evaluating postural risk
  factors associated with musculoskeletal disorders.
\newblock \emph{Applied Ergonomics}, 87:\penalty0 103138, 2020.

\bibitem[Li et~al.(2022)Li, Liu, Zhang, Xu, and Yan]{li2022cliff}
Zhihao Li, Jianzhuang Liu, Zhensong Zhang, Songcen Xu, and Youliang Yan.
\newblock {CLIFF}: Carrying location information in full frames into human pose
  and shape estimation.
\newblock In \emph{ECCV}, 2022.

\bibitem[Lin et~al.(2014)Lin, Maire, Belongie, Hays, Perona, Ramanan,
  Doll{\'a}r, and Zitnick]{lin2014coco}
Tsung-Yi Lin, Michael Maire, Serge Belongie, James Hays, Pietro Perona, Deva
  Ramanan, Piotr Doll{\'a}r, and C.~Lawrence Zitnick.
\newblock Microsoft {COCO}: Common objects in context.
\newblock In \emph{ECCV}, 2014.

\bibitem[Loper et~al.(2015)Loper, Mahmood, Romero, Pons-Moll, and
  Black]{loper2015smpl}
Matthew Loper, Naureen Mahmood, Javier Romero, Gerard Pons-Moll, and Michael~J.
  Black.
\newblock {SMPL}: A skinned multi-person linear model.
\newblock \emph{ACM Transactions on Graphics (Proc. SIGGRAPH Asia)},
  34\penalty0 (6), 2015.

\bibitem[Mehraban et~al.(2026)Mehraban, Iaboni, and Taati]{mehraban2026fasthmr}
Soroush Mehraban, Andrea Iaboni, and Babak Taati.
\newblock {FastHMR}: Accelerating human mesh recovery via token and layer
  merging.
\newblock In \emph{WACV}, 2026.

\bibitem[Mehta et~al.(2017)Mehta, Sridhar, Sotnychenko, Rhodin, Shafiei,
  Seidel, Xu, Casas, and Theobalt]{mehta2017vnect}
Dushyant Mehta, Srinath Sridhar, Oleksandr Sotnychenko, Helge Rhodin, Mohammad
  Shafiei, Hans-Peter Seidel, Weipeng Xu, Dan Casas, and Christian Theobalt.
\newblock {VNect}: Real-time {3D} human pose estimation with a single {RGB}
  camera.
\newblock \emph{ACM Transactions on Graphics}, 36\penalty0 (4):\penalty0
  44:1--44:14, 2017.

\bibitem[{NVIDIA}(2026)]{nvidia2026gemx}
{NVIDIA}.
\newblock {GEM-X}: Full-body and hand pose estimation from image and video.
\newblock \url{https://github.com/NVlabs/GEM-X}, 2026.

\bibitem[Patel and Black(2025)]{patel2025camerahmr}
Priyanka Patel and Michael~J. Black.
\newblock {CameraHMR}: Aligning people with perspective.
\newblock In \emph{3DV}, 2025.

\bibitem[Pavlakos et~al.(2019)Pavlakos, Choutas, Ghorbani, Bolkart, Osman,
  Tzionas, and Black]{pavlakos2019smplx}
Georgios Pavlakos, Vasileios Choutas, Nima Ghorbani, Timo Bolkart, Ahmed A.~A.
  Osman, Dimitrios Tzionas, and Michael~J. Black.
\newblock Expressive body capture: {3D} hands, face, and body from a single
  image.
\newblock In \emph{CVPR}, 2019.

\bibitem[{Qualcomm}(2026)]{qualcommaihub}
{Qualcomm}.
\newblock Qualcomm {AI Hub}.
\newblock \url{https://aihub.qualcomm.com}, 2026.

\bibitem[RangiLyu(2021)]{nanodet}
RangiLyu.
\newblock Nanodet-plus: Super fast and high accuracy lightweight anchor-free
  object detection model.
\newblock \url{https://github.com/RangiLyu/nanodet}, 2021.

\bibitem[Saito et~al.(2026)Saito, Li, de~Ruyter, Guerrero, Lim,
  et~al.]{saito2026soma}
Jun Saito, Jiefeng Li, Michael de Ruyter, Miguel Guerrero, Edy Lim, et~al.
\newblock {SOMA}: Unifying parametric human body models.
\newblock \emph{arXiv preprint arXiv:2603.16858}, 2026.

\bibitem[S{\'a}r{\'a}ndi and Pons-Moll(2024)]{sarandi2024nlf}
Istv{\'a}n S{\'a}r{\'a}ndi and Gerard Pons-Moll.
\newblock Neural localizer fields for continuous {3D} human pose and shape
  estimation.
\newblock In \emph{NeurIPS}, 2024.

\bibitem[{Seeing Machines}(2026)]{seeingmachines2026hmr}
{Seeing Machines}.
\newblock Seeing machines {HMR}: Real-time {3D} human mesh recovery at up to
  180 {FPS} on {NVIDIA Jetson Thor}.
\newblock Technical paper, \url{https://seeingmachines.com}, 2026.

\bibitem[Shen et~al.(2024)Shen, Pi, Xia, Cen, Peng, Hu, Bao, Hu, and
  Zhou]{shen2024gvhmr}
Zehong Shen, Huaijin Pi, Yan Xia, Zhi Cen, Sida Peng, Zechen Hu, Hujun Bao,
  Ruizhen Hu, and Xiaowei Zhou.
\newblock World-grounded human motion recovery via gravity-view coordinates.
\newblock In \emph{SIGGRAPH Asia}, 2024.

\bibitem[Shimada et~al.(2020)Shimada, Golyanik, Xu, and
  Theobalt]{shimada2020physcap}
Soshi Shimada, Vladislav Golyanik, Weipeng Xu, and Christian Theobalt.
\newblock {PhysCap}: Physically plausible monocular {3D} motion capture in real
  time.
\newblock \emph{ACM Transactions on Graphics}, 39\penalty0 (6):\penalty0 235,
  2020.

\bibitem[Shin et~al.(2024)Shin, Kim, Halilaj, and Black]{shin2024wham}
Soyong Shin, Juyong Kim, Eni Halilaj, and Michael~J. Black.
\newblock {WHAM}: Reconstructing world-grounded humans with accurate {3D}
  motion.
\newblock In \emph{CVPR}, 2024.

\bibitem[Song et~al.(2020)Song, Chen, and Hilliges]{song2020lgd}
Jie Song, Xu Chen, and Otmar Hilliges.
\newblock Human body model fitting by learned gradient descent.
\newblock In \emph{ECCV}, 2020.

\bibitem[Stathopoulos et~al.(2024)Stathopoulos, Han, and
  Metaxas]{stathopoulos2024scorehmr}
Anastasis Stathopoulos, Ligong Han, and Dimitris Metaxas.
\newblock Score-guided diffusion for {3D} human recovery.
\newblock In \emph{CVPR}, 2024.

\bibitem[Uhlrich et~al.(2023)Uhlrich, Falisse, Kidzi{\'n}ski, Muccini, Ko,
  Chaudhari, Hicks, and Delp]{uhlrich2023opencap}
Scott~D. Uhlrich, Antoine Falisse, {\L}ukasz Kidzi{\'n}ski, Julie Muccini,
  Michael Ko, Akshay~S. Chaudhari, Jennifer~L. Hicks, and Scott~L. Delp.
\newblock {OpenCap}: Human movement dynamics from smartphone videos.
\newblock \emph{PLOS Computational Biology}, 19\penalty0 (10):\penalty0
  e1011462, 2023.

\bibitem[von Marcard et~al.(2018)von Marcard, Henschel, Black, Rosenhahn, and
  Pons-Moll]{vonmarcard2018pw3d}
Timo von Marcard, Roberto Henschel, Michael~J. Black, Bodo Rosenhahn, and
  Gerard Pons-Moll.
\newblock Recovering accurate {3D} human pose in the wild using {IMUs} and a
  moving camera.
\newblock In \emph{ECCV}, 2018.

\bibitem[Wang and Daniilidis(2023)]{wang2023refit}
Yufu Wang and Kostas Daniilidis.
\newblock {ReFit}: Recurrent fitting network for {3D} human recovery.
\newblock In \emph{ICCV}, 2023.

\bibitem[Wang et~al.(2024)Wang, Wang, Liu, and Daniilidis]{wang2024tram}
Yufu Wang, Ziyun Wang, Lingjie Liu, and Kostas Daniilidis.
\newblock {TRAM}: Global trajectory and motion of {3D} humans from in-the-wild
  videos.
\newblock In \emph{ECCV}, 2024.

\bibitem[Wu et~al.(2017)]{wu2017aichallenger}
Jiahong Wu et~al.
\newblock {AI Challenger}: A large-scale dataset for going deeper in image
  understanding.
\newblock \emph{arXiv preprint arXiv:1711.06475}, 2017.

\bibitem[Yang et~al.(2026{\natexlab{a}})Yang, He, Jing, Yang, Liu, Zou, and
  Wang]{fastsam3dbody2026}
Timing Yang, Sicheng He, Hongyi Jing, Jiawei Yang, Zhijian Liu, Chuhang Zou,
  and Yue Wang.
\newblock Fast {SAM 3D Body}: Accelerating {SAM 3D Body} for real-time
  full-body human mesh recovery.
\newblock \emph{arXiv preprint arXiv:2603.15603}, 2026{\natexlab{a}}.

\bibitem[Yang et~al.(2026{\natexlab{b}})Yang, Kukreja, Pinkus, Sagar, Fan,
  Park, Shin, Cao, Liu, Ugrinovic, Feiszli, Malik, Doll{\'a}r, and
  Kitani]{yang2026sam3dbody}
Xitong Yang, Devansh Kukreja, Don Pinkus, Anushka Sagar, Taosha Fan, Jinhyung
  Park, Soyong Shin, Jinkun Cao, Jiawei Liu, Nicolas Ugrinovic, Matt Feiszli,
  Jitendra Malik, Piotr Doll{\'a}r, and Kris Kitani.
\newblock {SAM 3D Body}: Robust full-body human mesh recovery.
\newblock \emph{arXiv preprint arXiv:2602.15989}, 2026{\natexlab{b}}.

\bibitem[Zaoui et~al.(2026)Zaoui, B{\'e}gel, and Adjel]{zaoui2026rhythmic}
Yanis Zaoui, Valentin B{\'e}gel, and Mohamed Adjel.
\newblock Rhythmic fitness beat 'em all using real-time monocular human pose
  estimation.
\newblock In \emph{International Conference on Entertainment Computing (ICEC)},
  2026.

\bibitem[Zhang et~al.(2021)Zhang, Tian, Zhou, Ouyang, Liu, Wang, and
  Sun]{zhang2021pymaf}
Hongwen Zhang, Yating Tian, Xinchi Zhou, Wanli Ouyang, Yebin Liu, Limin Wang,
  and Zhenan Sun.
\newblock {PyMAF}: {3D} human pose and shape regression with pyramidal mesh
  alignment feedback loop.
\newblock In \emph{ICCV}, 2021.

\bibitem[Zhao et~al.(2024)Zhao, Lv, Xu, Wei, Wang, Dang, Liu, and
  Chen]{zhao2024rtdetr}
Yian Zhao, Wenyu Lv, Shangliang Xu, Jinman Wei, Guanzhong Wang, Qingqing Dang,
  Yi Liu, and Jie Chen.
\newblock {DETRs} beat {YOLOs} on real-time object detection.
\newblock In \emph{CVPR}, 2024.

\bibitem[Zhou et~al.(2019)Zhou, Barnes, Lu, Yang, and Li]{zhou2019continuity}
Yi Zhou, Connelly Barnes, Jingwan Lu, Jimei Yang, and Hao Li.
\newblock On the continuity of rotation representations in neural networks.
\newblock In \emph{CVPR}, 2019.

\end{thebibliography}
